\PassOptionsToPackage{table}{xcolor}
\documentclass{bmvc2k}

\usepackage{amssymb}
\usepackage{booktabs}
\usepackage{multirow}
\usepackage{wrapfig}

\title{ALSA: Anchors in Logit Space for Out-of-Distribution Accuracy Estimation}

\addauthor{Chenzhi Liu}{chenzhi.liu@uq.edu.au}{1}
\addauthor{Mahsa Baktashmotlagh}{m.baktashmotlagh@uq.edu.au}{1}
\addauthor{Yanran Tang}{yanran.tang@uq.edu.au}{1}
\addauthor{Zi Huang}{helen.huang@uq.edu.au}{1}
\addauthor{Ruihong Qiu}{r.qiu@uq.edu.au}{1}

\addinstitution{
 School of Electrical Engineering and Computer Science\\
 The University of Queensland\\
 Brisbane, Australia
}

\runninghead{LIU ET AL.}{ALSA}

\begin{document}

\maketitle

\begin{abstract}

Estimating model accuracy on unseen, unlabeled datasets is crucial for real-world machine learning applications, especially under distribution shifts that can degrade performance. Existing methods often rely on predicted class probabilities (softmax scores) or data similarity metrics. While softmax-based approaches benefit from representing predictions on the standard simplex, compressing logits into probabilities leads to information loss. Meanwhile, similarity-based methods can be computationally expensive and domain-specific, limiting their broader applicability. In this paper, we introduce ALSA (Anchors in Logit Space for Accuracy estimation), a novel framework that preserves richer information by operating directly in the logit space. Building on theoretical insights and empirical observations, we demonstrate that the aggregation and distribution of logits exhibit a strong correlation with the predictive performance of the model. To exploit this property, ALSA employs an anchor-based modeling strategy: multiple learnable anchors are initialized in logit space, each assigned an influence function that captures subtle variations in the logits. This allows ALSA to provide robust and accurate performance estimates across a wide range of distribution shifts. Extensive experiments on vision, language, and graph benchmarks demonstrate ALSA’s superiority over both softmax- and similarity-based baselines. Notably, ALSA’s robustness under significant distribution shifts highlights its potential as a practical tool for reliable model evaluation. Code has been released at \href{https://github.com/chenzhi-liu/ALSA}{https://github.com/chenzhi-liu/ALSA}.

\end{abstract}   
\section{Introduction}
\label{sec:intro}

In real-world machine learning model deployment, the trustworthy
and reliable performance of well-trained models largely depends on
the in-distribution (ID) assumption. However, the user interactions in online environments frequently introduce diverse and dynamically changing data streams, where the target (test) data may differ significantly from the source (train) data, resulting in substantial model performance degradation~\cite{recht2019imagenet, koh2021wilds, geirhos2018imagenet}. Consequently, accurately estimating model performance in these out-of-distribution (OOD) scenarios is of critical importance.

Performance estimation methods mainly fall into two categories. (1) Using the softmax scores of models. For instance, the Average Confidence (AC) method~\cite{guillory2021predicting} uses the average of maximum softmax scores to estimate the error, while Average Thresholded Confidence (ATC)~\cite{garg2022ATC} further refines the estimation by learning a threshold to better map softmax scores to error predictions. Similarly, Confidence Optimal Transport (COT)~\cite{lu2023characterizing} employs optimal transport between softmax scores under the assumption of a label distribution shift. (2) Relying on data similarity. These approaches suggest that a model’s accuracy is strongly associated with the similarity between the source and target datasets, and quantifying this similarity can lead to accurate performance estimation~\cite{deng2021labels, deng2021does}.

Although data similarity-based methods hold promise for model evaluation, they are computationally intensive, as their effectiveness often necessitates the use of multiple models or extensive data augmentation. Furthermore, these approaches are specifically designed for certain data types, including images~\cite{deng2021does,deng2021labels} and graphs~\cite{zheng2024gnnevaluator}, making them less suitable as a general framework for evaluating models under distribution shifts. Softmax-based methods are also limited as softmax compresses logits~\cite{goodfellow2016deep_ch6}, resulting in the loss of magnitude differences that are critical for accurate performance estimation. By operating in logit space instead of softmax outputs a greater amount of information is retained, thereby enhancing accuracy estimation.

Additionally, existing methods overlook class-wise imbalances, which can adversely impact predictions when training distributions are skewed~\cite{he2009learning, tang2020long}. For example, metrics like average confidence in AC~\cite{garg2022ATC} and thresholds in ATC~\cite{garg2022ATC} can be misleading when minority classes are scarce in training but prevalent in the target set. Thus, analyzing logits distribution and class-wise performance are both crucial for improving \textit{OOD performance estimation}.

This paper proposes \textit{Anchors in Logit Space for Accuracy estimation} (ALSA), a novel framework for estimating the accuracy of machine learning models on unlabeled datasets under distribution shift. Unlike existing methods that rely on softmax probabilities, ALSA operates directly in the logit space to preserve richer information for more accurate performance estimation. The approach is motivated by the observation that logits tend to aggregate near a specific hyperplane, and certain regions in logit space are more predictive of correct classification outcomes.

To formalize this insight, \textit{anchors} are introduced as reference points in logit space that act as confidence indicators. A probabilistic model is then constructed, where each anchor contributes to the likelihood that a logit vector corresponds to a correct prediction. These contributions are aggregated and passed through a sigmoid function to yield the final probability estimate. ALSA is evaluated on a diverse set of benchmarks, including vision, language, and graph datasets, and demonstrates improved accuracy estimation compared to existing methods. In addition to providing useful insights into model behavior under distribution shifts, it shows strong robustness in scenarios where the marginal distribution of the target variable differs across datasets, which is common in real-world applications.
\begin{itemize}
    \setlength{\itemsep}{-2pt}
    \setlength{\parsep}{0pt}
    \item Identification and analysis of information loss caused by softmax compression, emphasizing the importance of logit distribution in accuracy estimation.
    \item Development of ALSA, a novel framework that leverages logit-space distributions to estimate accuracy under distribution shifts without the need for labeled target data.
    \item Extensive evaluation across vision, language, and graph datasets, as well as diverse model architectures, demonstrating ALSA’s superior performance, robustness, and generalization ability in real-world applications.
\end{itemize}

\textbf{Prior Work.} Our research is related to but different from work in OOD detection~\cite{hendrycks2016baseline, liang2017enhancing, du2019implicit} and generalization estimation~\cite{recht2019imagenet, miller2021accuracy, garg2022ATC, baek2022agreement, zheng2024gnnevaluator}. Prior approaches use softmax scores~\cite{hendrycks2016baseline, garg2022ATC}, calibration~\cite{liang2017enhancing}, or dataset similarity~\cite{deng2021labels}, often at the cost of reduced generality or informativeness. ALSA addresses these issues by leveraging richer structures in logit space. Detailed related work can be found in Appendix~\ref{appendix:related_work}.
\section{Preliminaries and Motivations}
\label{sec:preliminaries}

\subsection{Problem Definition}

Consider a multi-class classification problem with input space $\mathcal{X} \in \mathbb{R}^d$ and label space $\mathcal{Y} = \{1, 2, \ldots, c\}$. The source and target datasets over $\mathcal{X} \times \mathcal{Y}$ are denoted by $D^S$ and $D^T$, with corresponding distributions $P_S$ and $P_T$. A classifier $f: \mathcal{X} \rightarrow \mathcal{Z}$ maps inputs to the logit space $\mathcal{Z} \in \mathbb{R}^c$, without softmax normalization. The training and validation sets $\{(\mathbf{x}_{\text{train}}^{(i)}, y_{\text{train}}^{(i)})\}$ and $\{(\mathbf{x}_{\text{val}}^{(i)}, y_{\text{val}}^{(i)})\}$ are drawn from $P_S$ and used to train and validate $f$. Given an unlabelled test set $D^T = \{\mathbf{x}_{\text{test}}^{(i)}\}_{i=1}^n$ sampled from a different distribution $P_T \neq P_S$, the goal is to estimate the accuracy of $f$ on $D^T$, denoted by $\text{Acc}(f, D^T)$.

\subsection{Band Width of Logit Distribution} 

Classification models generally follow a similar overarching architecture. The encoder \(\mathrm{En}(\cdot)\) transforms an input sample \(\mathbf{x}\) into an embedding vector \(\mathbf{h} \in \mathbb{R}^m\), which is then mapped to logits \(\mathbf{z} \in \mathbb{R}^c\) via a linear transformation. While different models vary in their encoder design, the logits they produce are constrained to a narrow band around a \((c-1)\)-dimensional hyperplane in the logit space. The extent of this fluctuation, termed band width, quantifies how far logits deviate from the hyperplane. The following proposition formalizes this constraint: 

\label{prop:prop1}
\textbf{Proposition 1.} \textit{Given a weight matrix \( W \in \mathbb{R}^{c \times m} \) with entries \( W_{ij} \sim N(0, \sigma_W^2) \), and an embedding vector \(\mathbf{h}\) whose components are independent with mean 0 and variance \(\sigma_h^2\), suppose \(\mathbf{b} = 0\) and Xavier initialization is applied, ensuring the variance condition of \(\mathbf{h}\). Then, the bandwidth of the logits satisfies:}
\(
\text{Bandwidth} < 4z \sqrt{\frac{1}{m+c}},
\)
\textit{where \( z \) is the z-score of a Gaussian distribution.}

The final linear transformation, given by \(\mathbf{z} = W\mathbf{h} + \mathbf{b}\), projects the embedding vector \(\mathbf{h}\) from an \(m\)-dimensional space to the \(c\)-dimensional logit space. These logits are then normalized by the softmax function to yield a probability vector \(\mathbf{z}'\) over the classes. A key observation is that the softmax function is invariant to constant shifts; that is, \(\mathrm{softmax}(\mathbf{z}) = \mathrm{softmax}(\mathbf{z} + t\mathbf{1})\) for any \(t \in \mathbb{R}\), where \(\mathbf{1} = (1, 1, \ldots, 1) \in \mathbb{R}^c\). This invariance implies that any gradient component along the direction of \(\mathbf{1}\) does not affect the output distribution \(\mathbf{z}'\). Consequently, during training with gradient-descent-based optimizers, updates to \(\mathbf{z}\) occur only in directions orthogonal to \(\mathbf{1}\), leaving its projection onto \(\mathbf{1}\) unchanged. As a result, the variation of the logits is confined to the \((c-1)\)-dimensional subspace orthogonal to \(\mathbf{1}\), effectively restricting them to lie within a narrow band around that hyperplane. The proposition above precisely quantifies the upper bound of this bandwidth. The formal proof is deferred to Appendix~\ref{appendix:proof_sec3}.

\begin{figure*}
\centering
\begin{tabular}{cccc}
\bmvaHangBox{
    \includegraphics[width=0.18\textwidth]{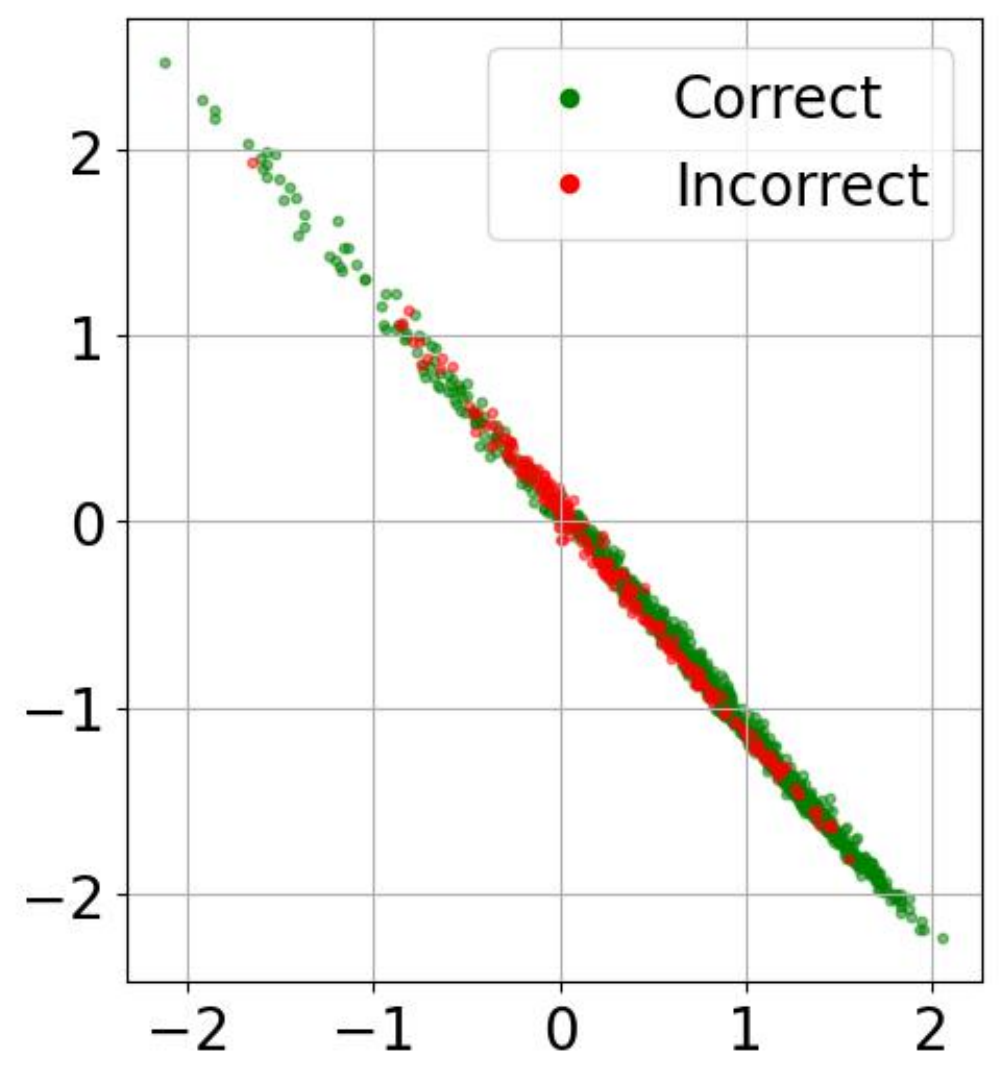}
} &
\bmvaHangBox{
    \includegraphics[width=0.18\textwidth]{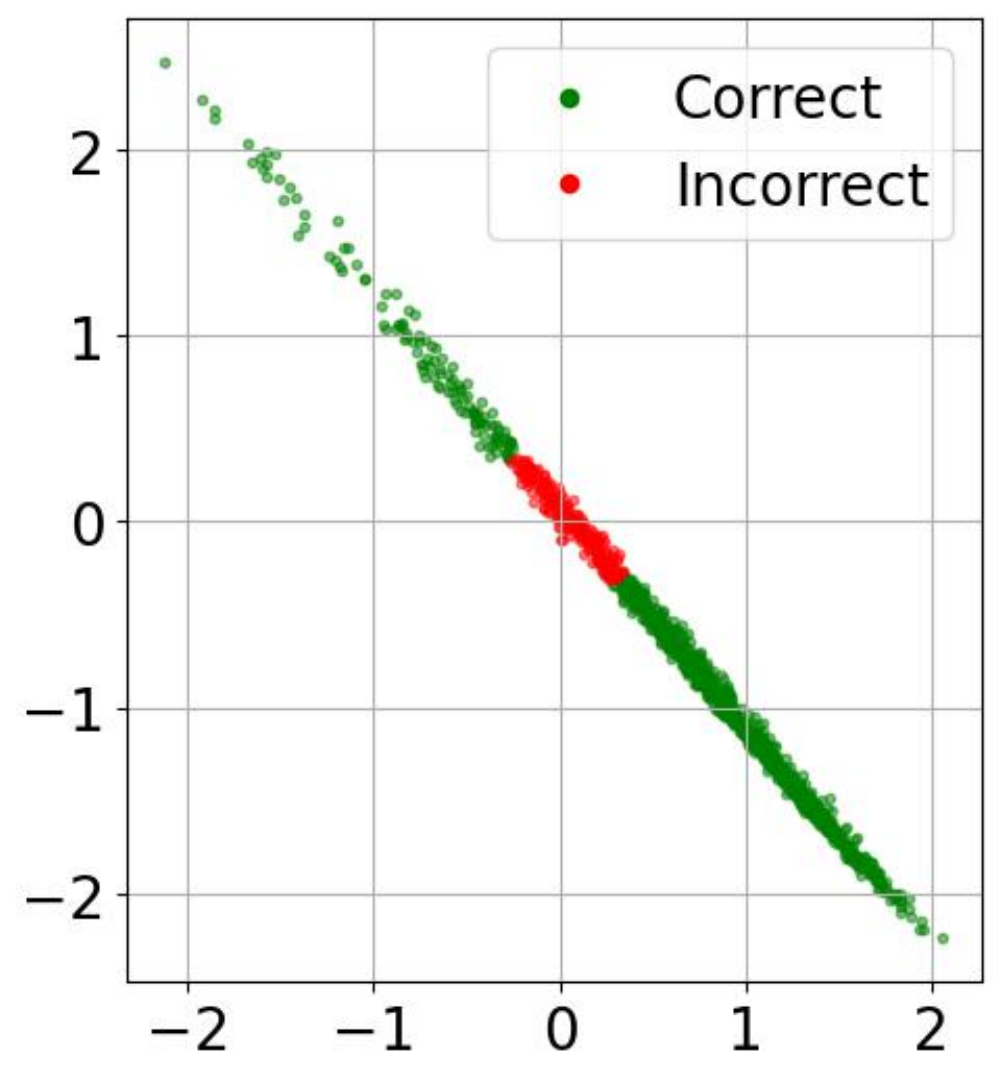}
} &
\bmvaHangBox{
    \includegraphics[width=0.24\textwidth]{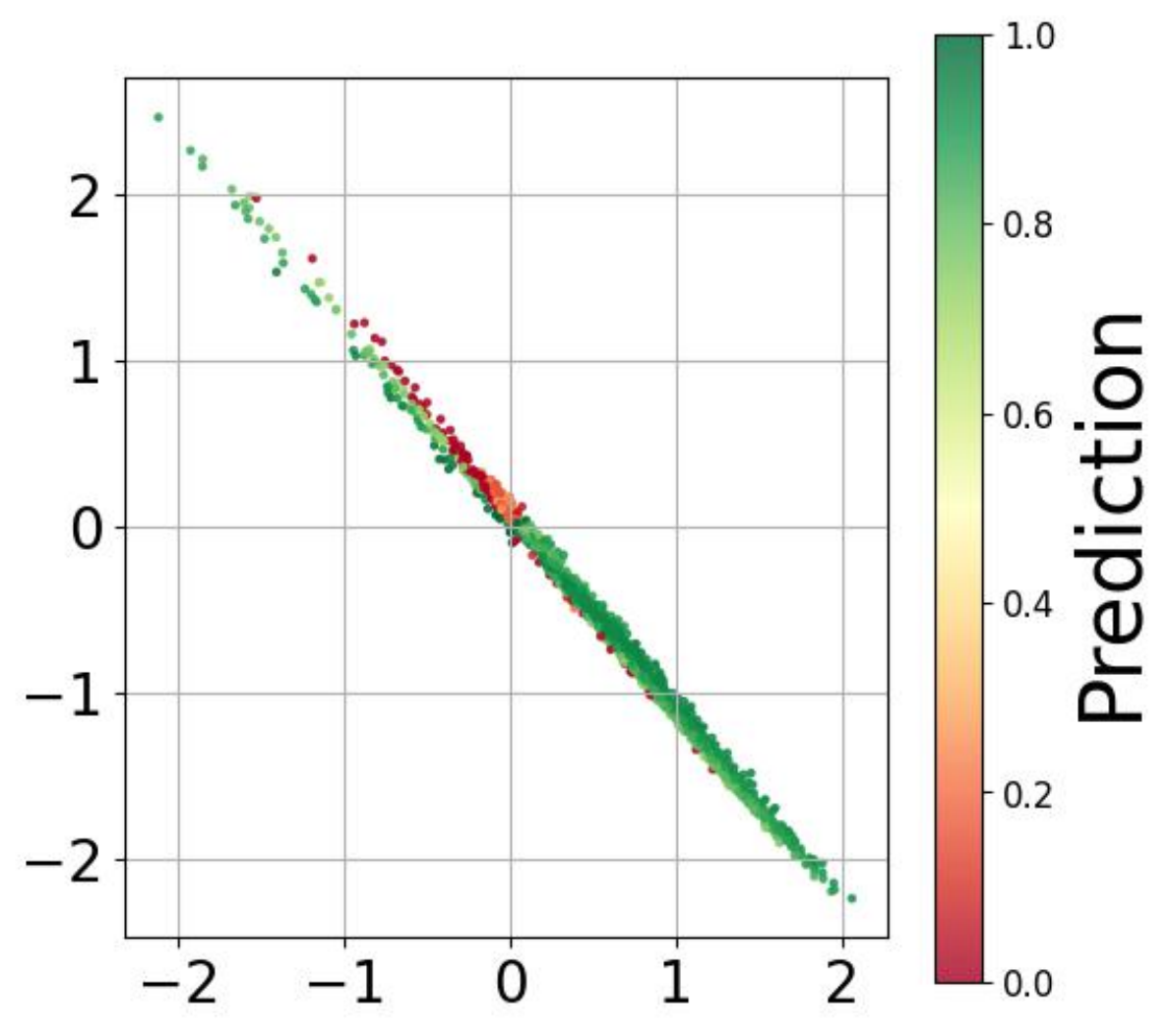}
} &
\bmvaHangBox{
    \includegraphics[width=0.24\textwidth]{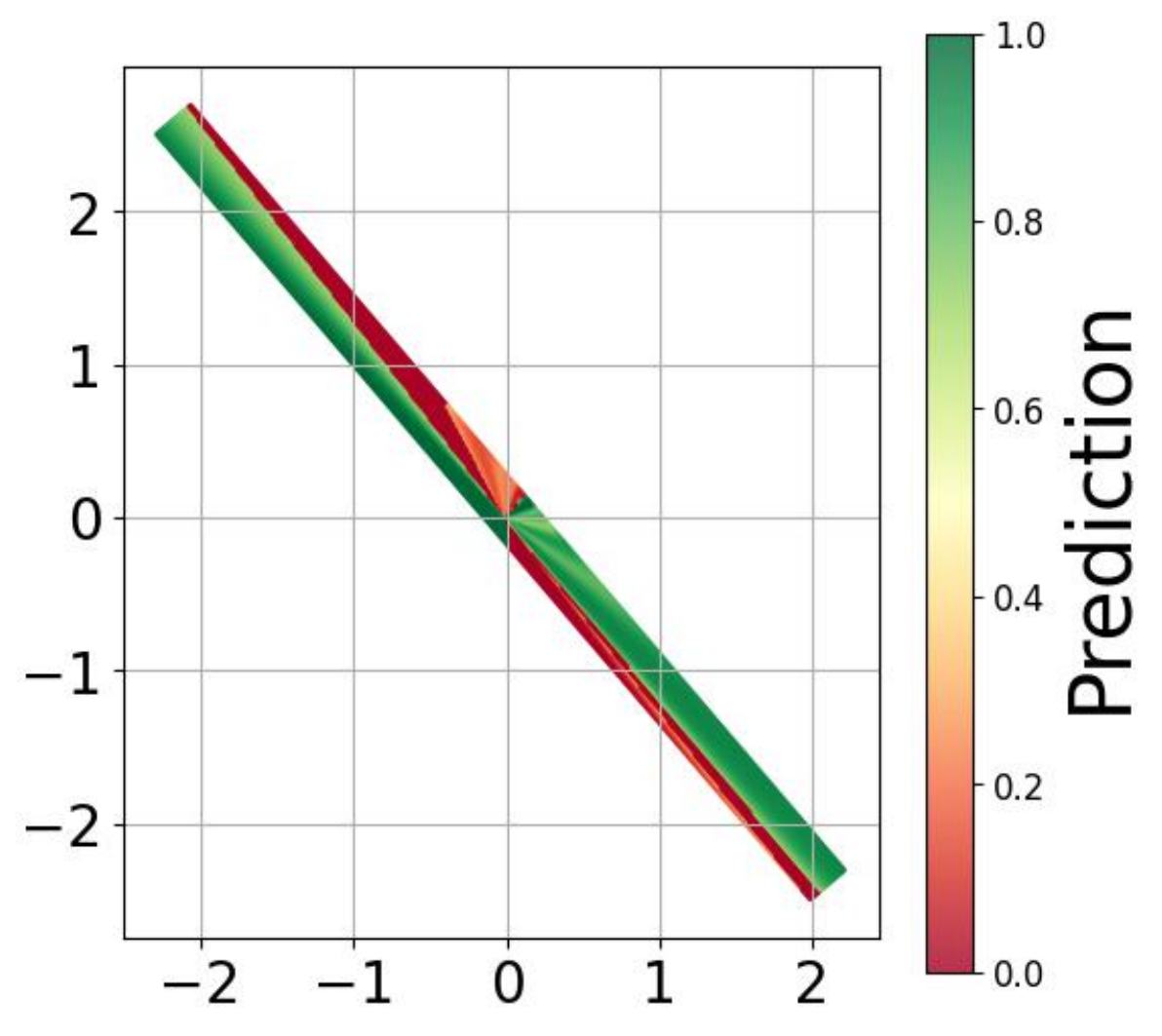}
} \\
(a) True pred. &
(b) ATC pred.  &
(c) ALSA pred. scatter &
(d) ALSA pred. surface 
\end{tabular}
\caption{Logit distributions and accuracy estimates on the Waterbirds-WILDS validation set using ResNet18. (a) Ground-truth correctness. (b) ATC predictions. (c) ALSA-estimated correctness probabilities. (d) ALSA’s prediction surface over the logit region. A 3D example for three classes is shown in Figure~\ref{fig:3d_vis} in appendix.}
\label{fig:2d_vis}
\end{figure*}

\subsection{Model Performance against Logit Distribution}

According to the proposition stated in Section \ref{prop:prop1}, the logits should closely follow a linear structure, as clearly illustrated in Figure~\ref{fig:2d_vis}. In Figure~\ref{fig:2d_vis}(a), each point represents a logit output, with colors indicating whether the predictions of a well-trained model are correct. The likelihood of correct predictions varies depending on the logit's position. For Class 0 (horizontal axis), incorrect samples (in red) are farther from the $y=x$ decision boundary, indicating that even with high confidence after softmax, the probability of incorrect predictions remains elevated. In contrast, for Class 1 (vertical axis), this uncertainty diminishes more rapidly, suggesting that Class 1 requires a lower confidence threshold for nearly perfect classification. This highlights the non-uniformity in the model’s ability to distinguish between different classes. Additionally, the distribution of incorrect samples suggests that logits carry information along the all-ones vector. For Class 0, incorrect samples are concentrated at the lower end of the band, whereas for Class 1, they are mostly at the upper end.

Figure~\ref{fig:2d_vis}(b) visualizes the predictions of ATC, which applies a unified softmax score threshold to classify all samples below it as incorrect. However, this approach overlooks variations in the probability of correct predictions across different regions in logit space and fails to account for class-wise differences in classification difficulty.

A similar trend holds in a three-class synthetic scenario (detailed in Appendix \ref{appendix:logits_distribution_3d}), where logits cluster near a plane in 3D space. This further supports the hypothesis that the position of logits provides additional insight into the probability of correct predictions. To explore this, ALSA models logits' structure and its relationship with correctness.
\section{Methodology}

\subsection{Definition of Anchors}

Since all the logits aggregate close to a hyperplane perpendicular to the all-ones vector, it can be expected that when a vector is far from the all-ones vector, certain elements of the vector will dominate. Such a vector will have high confidence after normalization by the softmax function. However, a high-confidence predicted probability does not always indicate a correct prediction. Additionally, the probability of a correct prediction is expected to decay when no element in the predicted probability dominates, indicating that the model is more uncertain among possible classes. Therefore, logits at different positions on the hyperplane correspond to different probabilities of the model making a correct prediction. A set of predefined anchors is used to capture this variation in probability.

An \textit{anchor} is defined as a triplet consisting of a position vector \( \mathbf{a} \in \mathbb{R}^c \), a peak influence value \( p \in \mathbb{R} \), and a variance \( v \in \mathbb{R} \). The set of \( k \) anchors is denoted by \( \mathcal{A} = \{ (\mathbf{a}_i, p_i, v_i) \}_{i=1}^k \), where $\mathrm{a}_i$, $p_i$, and $v_i$ are three learnable parameters. Each anchor exerts influence on other logits, and the cumulative influences are transformed into a probability of correct prediction using a sigmoid function. For logits with high positive influences, a high probability of correct prediction is expected. Similarly, for those with negative influences, it is highly probable they will result in incorrect predictions.

\noindent\textbf{Anchors Initialization.} The $k$ anchor positions are initialized by randomly sampling $k$ logits from the validation set. Peak values are set to a positive or negative constant, depending on whether the corresponding logit represents a correct or incorrect prediction. Variances are drawn from a Gaussian distribution with moderate mean and variance to allow sufficient flexibility. ALSA is not sensitive to the specific choice of initialization, as its objective is to learn anchors that align with the underlying logit distribution. Since anchor parameters are continuously refined during training, different initializations tend to converge to similar distribution given the same data. A comparison between random initialization and sampling anchor positions from the validation set, presented in Appendix~\ref{appendix:init_method}, shows nearly identical prediction surfaces, supporting the robustness of ALSA to initialization choices.

\begin{wrapfigure}{r}{0.47\textwidth}
    \centering
    \vspace{-10pt} 
    \includegraphics[width=0.47\textwidth]{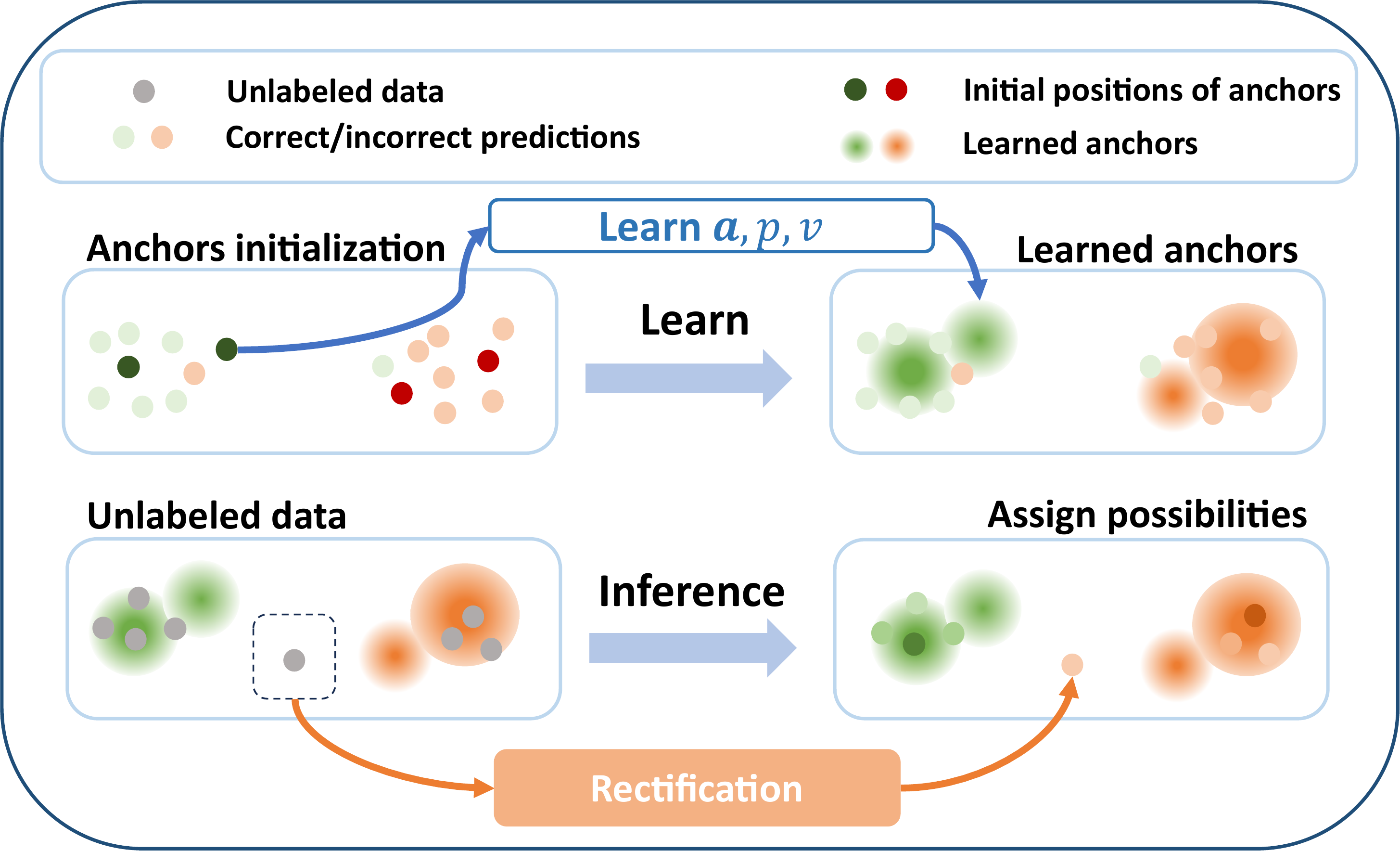}
    \caption{The workflow of ALSA: Anchors are learned from the validation set. During the testing phase, the probability of correctness is assigned to each logit. For regions where the anchor cannot generalize well, rectification is performed.}
    \label{fig:pipeline}
    \vspace{-12pt} 
\end{wrapfigure}

\subsection{Influence Function of Anchors}

Let the logit vector $\mathbf{z}_i$ from the classifier $f$ correspond to input sample $\mathbf{x}_i$, that is, $\mathbf{z}_i = f(\mathbf{x}_i)$. 

In order to quantify the influence that a set of anchors $\mathcal{A}$ exerts on the logit vector $\mathbf{z}_i$, an influence function $\mathrm{Infl}(\cdot,\cdot)$ is required. Ideally, this function should:
\begin{itemize}
    \setlength{\itemsep}{-2pt}
    \setlength{\parsep}{0pt}
    \item Exert an influence that diminishes as the distance between the logit vector and the anchor increases.
    \item Have parameters to control the maximum influence and distance an anchor can exert.
    \item Be differentiable to facilitate optimization.
\end{itemize}

Given these criteria, a Gaussian-like influence function is proposed for a specific anchor $(\mathbf{a}_j, p_j, v_j)$:
\begin{equation}
    \mathrm{Infl}(\mathbf{z}_i, (\mathbf{a}_j, p_j, v_j)) = p_j \exp\left(-v_j^2 \, \mathrm{dist}^2(\mathbf{z}_i, \mathbf{a}_j)\right).
\end{equation}
The magnitude $|p_j|$ controls the maximum influence an anchor can exert on a logit, and its sign determines the nature of the influence (positive or negative). As the squared distance $\mathrm{dist}^2(\mathbf{z}_i, \mathbf{a}_j)$ increases, the influence decreases rapidly. The variance $v_j$ controls the decay rate of the influence; a smaller $v_j$ value allows the influence to propagate over a greater distance.

The total influence that a logit $\mathbf{z}_i$ receives from the set of anchors $\mathcal{A}$ is defined as:
\begin{equation}
    \mathrm{Infl}(\mathbf{z}_i, \mathcal{A}) = \sum_{j=1}^k p_j \exp\left(-v_j^2 \, \mathrm{dist}^2(\mathbf{z}_i, \mathbf{a}_j)\right),
\end{equation}
where the distance function $\mathrm{dist}$ is the cosine distance between two vectors, defined as
\begin{equation}
    \mathrm{dist}(\mathbf{z}_i, \mathbf{a}_j) = 1 - \frac{\mathbf{z}_i \cdot \mathbf{a}_j}{\|\mathbf{z}_i\| \|\mathbf{a}_j\|}.
\end{equation}

The bounded nature of cosine distance within [0, 2] helps maintain parameter stability by preventing extreme values, thereby promoting a more stable training. Finally, the probability that the model will give a correct prediction when the logits are $\mathbf{z}_i$ is given by:
\begin{equation}
p_{\text{true}}(\mathbf{z}_i) = \mathrm{Sigmoid}\left(\mathrm{Infl}(\mathbf{z}_i, \mathcal{A})\right).
\end{equation}
Although the Gaussian-like function satisfies the desired properties, the choice of influence function is not unique. An alternative approach is to use an exponential influence function, which is defined as:
\begin{equation}
    \mathrm{Infl}(\mathbf{z}_i, (\mathbf{a}_j, p_j, v_j)) = p_j \exp\left(-v_j^2 \, \mathrm{dist}(\mathbf{z}_i, \mathbf{a}_j)\right).
\end{equation}
Unlike the Gaussian-like function, which decays quadratically with distance, the exponential function decreases more rapidly when the distance is small but more gradually when the distance is large. To examine its effect, this alternative formulation is also evaluated in the experiments.

\subsection{Model Inference}
For an unlabeled target dataset with $n$ samples $D^T = \{ \mathbf{x}_i \}_{i=1}^n$, the corresponding logits set is denoted as $S^T = \{ \mathbf{z}_i \}_{i=1}^n$. The initial predicted accuracy of the evaluated model $f$ on target dataset $D^T$ is provided as the arithmetic mean of the probability of correct prediction for all samples:
\begin{equation}
\hat{Acc}_{\text{train}}(f, D^T) = \frac{1}{n} \sum_{i=1}^n p_{\text{true}}(\mathbf{z}_i).
\end{equation}

\textbf{Estimation Rectification.} Some logits may receive minimal influence from all anchors, indicating they are located far from known regions in logit space. In such cases, the total influence $\mathrm{Infl}(\mathbf{z}_i, \mathcal{A})$ approaches zero, leading to an estimated probability near 50\% after applying the sigmoid function. However, this is unreliable, as it reflects high uncertainty rather than an informed prediction. To address this, a simple rectification is applied: if a sample's logits receive insufficient influence, the prediction is set to $p_{\text{true}}(\mathbf{z}_i) = \frac{1}{c}$, where $c$ is the number of classes.

A logit is deemed negligibly influenced if it falls outside a pre-specified confidence interval of an anchor’s influence peak, controlled by a confidence interval $\alpha \in (0,1)$. Specifically, for a Gaussian-like influence function, the cut-off is given by $t = pe^{-\left( \mathrm{erf}^{-1}(\alpha) \right)^2}$, where $\mathrm{erf}^{-1}(\cdot)$ is the inverse error function, and for an exponential influence function, it is $t=p(1-\alpha)$. This cut-off, derived from the confidence interval $\alpha$, is independent of variance, serving as a unified threshold for all anchor influences in the two proposed influence functions. Proof is in Appendix~\ref{appendix:anchors_t}.

Finally, the predicted accuracy is:
\begin{equation}
    \hat{Acc}(f, D^T) = \frac{1}{n} \sum_{i=1}^n p_{\text{rec}}(\mathbf{z}_i),
\end{equation}
where $p_{\text{rec}}(\mathbf{z}_i)$ is defined as:
\begin{equation}
\begin{cases} 
\mathrm{Sigmoid}\left( \mathrm{Infl}(\mathbf{z}_i, \mathcal{A}) \right) & \left| \mathrm{Infl}(\mathbf{z}_i, \mathcal{A}) \right| \geq t, \\
\frac{1}{c} & \text{otherwise},
\end{cases}
\end{equation}
where \(c\) is the number of classes.

\subsection{Learning Objective}

The source validation set is denoted as $D = \{ (\mathbf{x}_{\text{val}}^{(i)}, y_{\text{val}}^{(i)}) \}_{i=1}^n$ with $n$ samples. Using the classifier $f$, the logits for each sample are obtained, resulting in the set $\{ (\mathbf{z}_{\text{val}}^{(i)}, y_{\text{val}}^{(i)}) \}_{i=1}^n$, where $\mathbf{z}_{\text{val}}^{(i)} = f(\mathbf{x}_{\text{val}}^{(i)})$. The predicted label is defined as $\hat{y}_i = \arg\max_j z_{\text{val},j}^{(i)}$, where $z_{\text{val},j}^{(i)}$ is the $j$-th element of $\mathbf{z}_{\text{val}}^{(i)}$. Additionally, $\delta_i = \mathbb{I}(\hat{y}_i = y_{\text{val}}^{(i)})$ is defined. Thus, the set $\{ (\mathbf{z}_{\text{val}}^{(i)}, \delta_i) \}_{i=1}^n$ is obtained. The set of anchors with $k$ anchors is $\mathcal{A} = \{ (\mathbf{a}_j, p_j, v_j) \}_{j=1}^k$.

For each $\mathbf{z}_{\text{val}}^{(i)}$, $p_{\text{true}}(\mathbf{z}_{\text{val}}^{(i)})$ is defined as the probability that the model will make a correct prediction. The Binary Cross Entropy Loss is used as the loss function. By minimizing
\begin{equation}
\begin{aligned}
\mathcal{L} = - \sum_{i=1}^n \Big[ & \delta_i \cdot \log\left( p_{\text{true}}(\mathbf{z}_{\text{val}}^{(i)}) \right) + (1 - \delta_i) \cdot \log\left( 1 - p_{\text{true}}(\mathbf{z}_{\text{val}}^{(i)}) \right) \Big],
\end{aligned}
\end{equation}

where the set of anchors can be learned that reflects the distribution of logits and the probability that the model will give correct predictions when logits are in different regions.

The stopping criteria for the training process must also be considered. The model's true accuracy on $D$ is given by $Acc(f, D) = \frac{1}{n} \sum_{i=1}^n \delta_i$. After each training epoch, the predicted accuracy is calculated as $\hat{Acc}_{\text{train}}(f, D) = \frac{1}{n} \sum_{i=1}^n p_{\text{true}}(\mathbf{z}_{\text{val}}^{(i)})$. The training is stopped when
\begin{equation}
\left| \hat{Acc}_{\text{train}}(f, D) - Acc(f, D) \right| < \epsilon.
\end{equation}
Here, $\epsilon$ is a very small positive number in the range $(0, 1]$; in practice, $\epsilon = 10^{-5}$ is used for all experiments.

\section{Experiments}

\textbf{Datasets.} Multiple benchmark datasets across different types of distributions, including vision, language, and graph datasets, are considered in this study. \textbf{Natural shift:} Non-simulated shifts from data collection variations are evaluated using CIFAR-10.1~\cite{recht2018cifar101}, CIFAR-10.2~\cite{lu2020harder}, and ImageNetV2~\cite{recht2019imagenet}. \textbf{Subpopulation shift:} Subpopulation changes are examined using Waterbirds-WILDS (W-WILDS)~\cite{sagawa2019distributionally} for vision tasks and Amazon-WILDS (A-WILDS)~\cite{wilds2021} for language tasks. \textbf{Synthetic shift:} Deliberate corruptions are tested using CIFAR-10C~\cite{hendrycks2019benchmarking}, CIFAR-100C~\cite{hendrycks2019benchmarking}, and MNIST-M~\cite{zhao2020review}. \textbf{Domain shift:} Domain adaptation is evaluated using the Office-31 dataset~\cite{koniusz2017domain}, with data from three domains. \textbf{Temporal shift:} Temporal changes are assessed using the ogbn-arxiv dataset~\cite{hu2020ogb}, focusing on time-evolving data. Further dataset setup details are in Appendix~\ref{appendix:dataset_details}.
\vspace{0.2\baselineskip}

\noindent\textbf{Baselines.} The proposed method, ALSA, is compared with the baseline methods. \textbf{Average Confidence (AC)}~\cite{garg2022ATC} estimates error by averaging maximum softmax confidence. \textbf{Difference of Confidence (DoC)}~\cite{guillory2021predicting} uses the difference in confidence between source and target data. \textbf{Importance Re-weighting (IM)} re-weights classifier error using confidence-based slices~\cite{chen2021mandoline}. \textbf{Average Thresholded Confidence (ATC)}~\cite{garg2022ATC} estimates error by thresholding softmax confidence values. \textbf{Confidence Optimal Transport (COT)}~\cite{lu2023characterizing} uses optimal transport cost to quantify distributional shifts. \textbf{GNNEvaluator (GNNEval)}~\cite{zheng2024gnnevaluator} predicts GNN performance using simulated graph data. ALSA with an exponential influence function is denoted as ALSA-E, while ALSA with a Gaussian-like influence function is denoted as ALSA-G or simply ALSA. Detailed explanations for baselines are in Appendix~\ref{appendix:baseline_details} and the details of experimental setup can be found in Appendix ~\ref{appendix:exp_setup}.

\subsection{Results}

\begin{table*}
\centering
\begin{minipage}[t]{0.48\textwidth}
\centering
\caption{MAE of accuracy estimates on vision and language datasets, averaged over three seeds. Bold: best; Italics: second-best among ours. Full table with std in Appendix~\ref{appendix:exp_details}.}
\label{tab:mae_cv_language}
\renewcommand{\arraystretch}{1.3}
\scalebox{0.45}{
\begin{tabular}{llccccc>{\columncolor{gray!10}}c>{\columncolor{gray!10}}c}
\toprule
\textbf{Dataset} & \textbf{Variant} & \textbf{AC} & \textbf{DoC} & \textbf{IM} & \textbf{ATC} & \textbf{COT} & \textbf{ALSA-E} & \textbf{ALSA-G} \\
\midrule
\multirow{3}{*}{CIFAR10} 
 & CIFAR-10.1   & 4.82 & 4.69 & 5.70 & 1.27 & 3.02 & 2.83 & \textbf{1.22} \\
 & CIFAR-10.2   & 7.97 & 7.84 & 8.74 & 4.59 & 6.20 & \textbf{0.61} & \textit{4.19} \\
 & CIFAR-10C    & 6.57 & 6.44 & 7.60 & \textbf{2.28} & 3.18 & 5.09 & 4.18 \\
\midrule
CIFAR100 & CIFAR-100C & 9.46 & 7.80 & 9.02 & 3.85 & \textbf{2.41} & 7.26 & 6.56 \\
\midrule
\multirow{2}{*}{A-WILD} 
 & Subpop1 & 3.70 & 2.16 & 2.14 & 5.90 & 3.39 & \textit{0.99} & \textbf{0.97} \\
 & Subpop2 & 4.19 & 2.65 & 2.65 & 5.82 & 4.18 & \textit{1.43} & \textbf{1.17} \\
\midrule
W-WILDS & Subpop & 0.78 & 0.79 & 0.74 & 1.55 & 0.97 & \textbf{0.32} & \textit{0.54} \\
\midrule
ImageNet & ImageNetv2 & 3.63 & 2.68 & 2.85 & \textbf{0.75} & 9.57 & 4.07 & 3.22 \\
\midrule
MNIST & MNIST-M & 13.27 & 13.44 & 16.29 & 13.63 & 4.96 & \textbf{2.82} & \textit{3.09} \\
\midrule
Office-31 & domain & 38.67 & 34.47 & 32.90 & 41.55 & 11.40 & \textit{8.35} & \textbf{7.77} \\
\bottomrule
\end{tabular}
}
\end{minipage}
\hfill
\begin{minipage}[t]{0.49\textwidth}
\centering
\caption{MAE between estimated and true accuracy on ogb-arxiv, averaged over three seeds. AVG. denotes the mean across backbones. Full results with standard deviations are in Appendix~\ref{appendix:exp_details}.}
\label{tab:mae_gnn}
\renewcommand{\arraystretch}{1.3}
\scalebox{0.5}{
\begin{tabular}{llccccc>{\columncolor{gray!10}}c>{\columncolor{gray!10}}c}
\toprule
\textbf{Model} & \textbf{AC} & \textbf{DoC} & \textbf{IM} & \textbf{GNNEval} & \textbf{ATC} & \textbf{COT} & \textbf{ALSA-E} & \textbf{ALSA-G} \\
\midrule
GCN     & 1.61 & 2.94 & 3.89 & 10.59 & 19.76 & 6.64 & \textit{1.13} & \textbf{0.99} \\
SAGE    & 1.31 & 1.72 & 2.07 & 10.42 & 25.31 & 5.85 & 2.21 & \textbf{1.08} \\
GAT     & 3.24 & 4.16 & 4.60 & 13.34 & 23.89 & 5.06 & \textit{1.73} & \textbf{1.01} \\
GIN     & 30.09 & 5.86 & 7.17 & 9.54 & 32.28 & 30.79 & \textit{2.18} & \textbf{2.14} \\
\midrule
AVG.    & 9.06 & 3.67 & 4.43 & 10.97 & 25.31 & 12.09 & \textit{1.81} & \textbf{1.30} \\
\bottomrule
\end{tabular}
}
\end{minipage}
\end{table*}

\noindent\textbf{Results across Modalities.} Tables~\ref{tab:mae_cv_language} and \ref{tab:mae_gnn} summarize MAE results on vision/language and graph datasets, respectively. The proposed method consistently ranks among the top performers across diverse distribution shifts. It significantly outperforms all baselines on subpopulation benchmarks (e.g., Amazon-WILDS, Waterbirds-WILDS) and achieves strong results under natural shifts (e.g., CIFAR10.1, CIFAR10.2). On synthetic corruptions (e.g., CIFAR-10C, CIFAR-100C), it remains robust across a wide severity range, with Appendix~\ref{appendix:exp_details} showing accurate estimates even when true accuracy drops by over 30\%. As no method can guarantee correctness under all conditions due to the nature of the task~\cite{garg2022ATC}, it is notable that while baselines exceed 10\% error on at least one dataset, our method consistently remains below this threshold. In the graph domain (ogbn-arxiv), it delivers clear gains across all GNN backbones under temporal shifts.

\noindent\textbf{Correlation with True Accuracy.}
Table~\ref{tab:gnn_performance_metrics} reports $R^2$ and Pearson correlation ($\rho$) values on ogbn-arxiv across years. The proposed method consistently achieves the high correlation scores across models, demonstrating its ability to track true performance under distribution shift.

\begin{wraptable}[11]{r}{0.62\textwidth}
\vspace{-5pt}
\caption{$R^2$ and Pearson correlation ($\rho$) between estimated and true accuracy. $R^2$ reflects fit to $y = x$; $\rho$ measures linear correlation. Higher is better. Best results are highlighted.}
\label{tab:gnn_performance_metrics}
\centering
\renewcommand{\arraystretch}{1.1}
\setlength{\tabcolsep}{2pt}
\tiny
\begin{tabular}{lccccccccccccccc}
\toprule
\multirow{2}{*}{\textbf{Model}} & \multicolumn{2}{c}{\textbf{AC}} & \multicolumn{2}{c}{\textbf{DoC}} & \multicolumn{2}{c}{\textbf{IM}} & \multicolumn{2}{c}{\textbf{GNNEval}} & \multicolumn{2}{c}{\textbf{ATC}} & \multicolumn{2}{c}{\textbf{COT}} & \multicolumn{2}{c}{\textbf{ALSA}} \\
\cmidrule(lr){2-3} \cmidrule(lr){4-5} \cmidrule(lr){6-7} \cmidrule(lr){8-9} \cmidrule(lr){10-11} \cmidrule(lr){12-13} \cmidrule(lr){14-15}
& $R^2$ & $\rho$ & $R^2$ & $\rho$ & $R^2$ & $\rho$ & $R^2$ & $\rho$ & $R^2$ & $\rho$ & $R^2$ & $\rho$ & $R^2$ & $\rho$ \\
\midrule
GCN   & 0.84 & 0.96 & 0.52 & 0.96 & 0.25 & 0.96 & -29.28 & -0.23 & -17.42 & 0.95 & -1.12 & 0.96 & \textbf{0.94} & \textbf{0.97} \\
SAGE  & 0.85 & 0.97 & 0.74 & 0.98 & 0.61 & 0.98 & -29.10 & -0.29 & -48.32 & \textbf{0.98} & -2.13 & 0.96 & \textbf{0.87} & \textbf{0.98} \\
GAT   & -0.13 & 0.85 & -0.50 & 0.84 & -0.73 & 0.85 & -44.46 & -0.25 & -29.58 & 0.55 & -0.74 & 0.75 & \textbf{0.91} & \textbf{0.99} \\
GIN   & -162.85 & 0.85 & -6.40 & \textbf{0.85} & -8.69 & 0.76 & -19.58 & -0.16 & -189.22 & 0.82 & -170.22 & 0.83 & \textbf{-0.32} & 0.46 \\
\midrule
AVG.  & -40.32 & 0.91 & -1.41 & 0.91 & -2.14 & 0.89 & -30.60 & -0.23 & -71.13 & 0.83 & -43.55 & 0.88 & \textbf{0.60} & \textbf{0.85} \\
\bottomrule
\end{tabular}
\vspace{-5pt}
\end{wraptable}

\noindent Additional analyses are provided in the appendix to support the robustness and interpretability of ALSA. Appendix~\ref{appendix:hyperparameters_analysis} examines the effects of key hyperparameters, including the confidence interval $\alpha$ and the number of anchors. The analysis shows that while extreme values of the confidence interval $\alpha$ can lead to increased estimation error, ALSA remains robust across a broad and practical range of settings. Performance also stabilizes once a sufficient number of anchors is reached. Appendix~\ref{appendix:complexity_analysis} presents a complexity analysis, showing that inference time scales linearly with the number of samples. Inference and training time comparisons are also reported, demonstrating a favorable trade-off between accuracy and efficiency.

\begin{figure}[!h]
\centering
\begin{tabular}{cc}
\bmvaHangBox{
    \includegraphics[width=0.34\columnwidth]{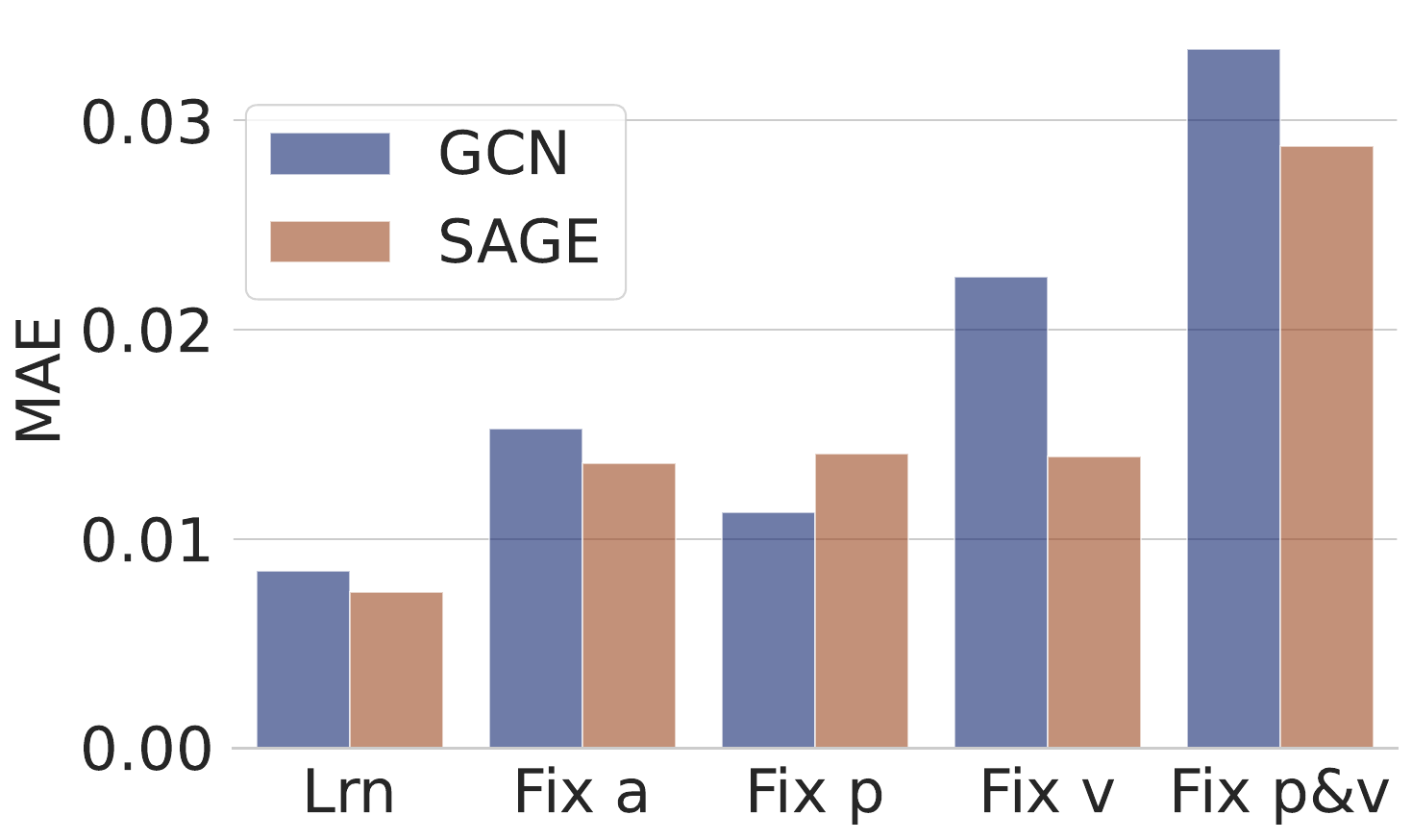}
} &
\bmvaHangBox{
    \includegraphics[width=0.37\columnwidth]{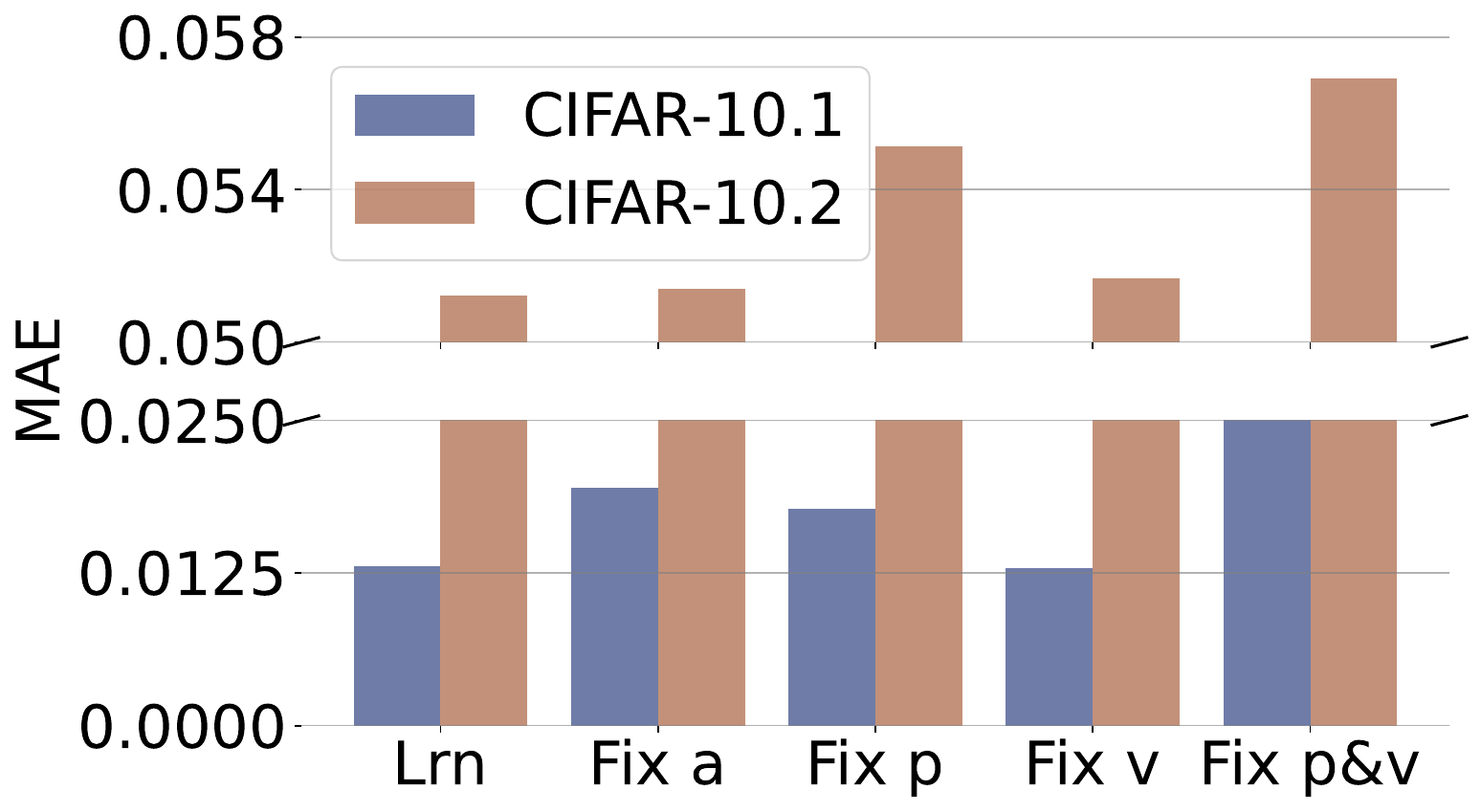}
} \\
(a) GCN and SAGE for ogbn-arxiv & (b) CIFAR-10.1 and CIFAR-10.2
\end{tabular}
\caption{(a) MAE is reported to evaluate the GCN and SAGE models on the ogbn-arxiv dataset by ablating different modules. (b) MAE is reported to evaluate the ResNet18 model on CIFAR-10.1 and CIFAR-10.2 by ablating different modules.}
\label{fig:ablation}
\end{figure}

\subsection{Ablation Study}
The ablation study evaluated the effectiveness of the learnable parameters in ALSA with 1.\ anchor position ($\mathrm{a}$), representing each anchor's position in logit space; 2.\ peak influence value ($p$), indicating the maximum positive or minimum negative influence an anchor can exert; and 3.\ variance ($v$), determining the extent of the influence. To test different configurations, these parameters are either fixed after initialization or updated during training. Five experiments are conducted: 1.\ \texttt{Lrn}: all parameters ($\mathrm{a}$, $p$, and $v$) are learnable; 2.\ \texttt{Fix a}: anchor positions are fixed; 3.\ \texttt{Fix p}: peak values are fixed; 4.\ \texttt{Fix v}: variances are fixed; and 5.\ \texttt{Fix p\&v}: both $p$ and $v$ are fixed.

MAE is utilised as the metric of ablation study. As shown in Figure~\ref{fig:ablation}, ablating any component leads to an increase in error, whereas the fully learnable configuration consistently attains the lowest error. Although the severity of the degradation varies, the learnable configuration consistently outperforms the others. The anchor position defines the region being modeled, and making it non-learnable (\texttt{Fix a}) restricts the model to focusing on the probability distribution of only specific regions. Fixing any of the peak influence value (\texttt{Fix p}), the variance (\texttt{Fix v}), and both (\texttt{Fix p\&v}) hinders the anchors to provide a more fine-grained influence decay, which results in worse generalization of probability distribution.

\subsection{Visualization of Anchors}
\label{appendix:vis_perd_anchors}

\begin{figure}[!t]
\centering
\begin{tabular}{cccc}
\bmvaHangBox{
    \includegraphics[width=0.225\columnwidth]{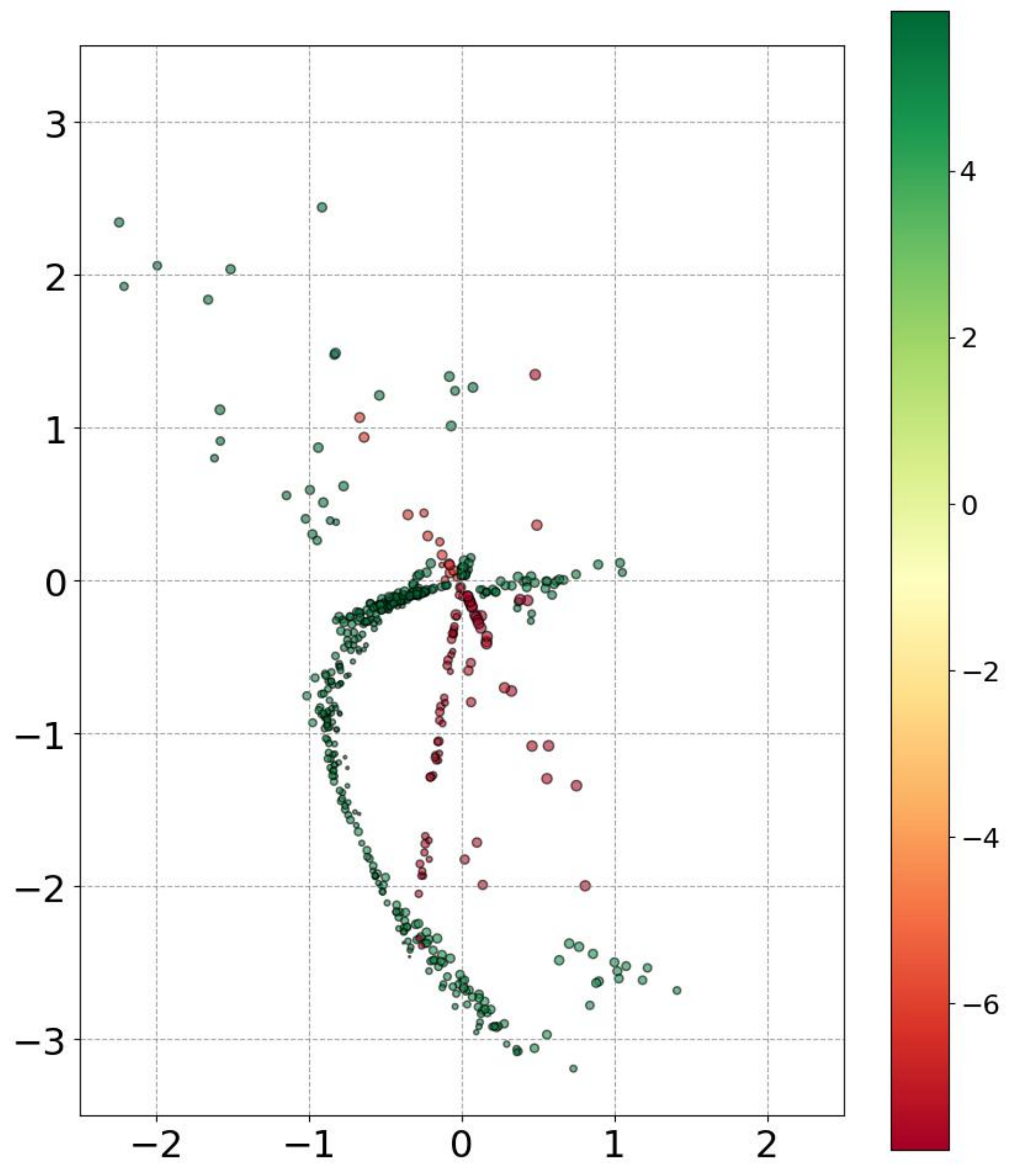}
} &
\bmvaHangBox{
    \includegraphics[width=0.225\columnwidth]{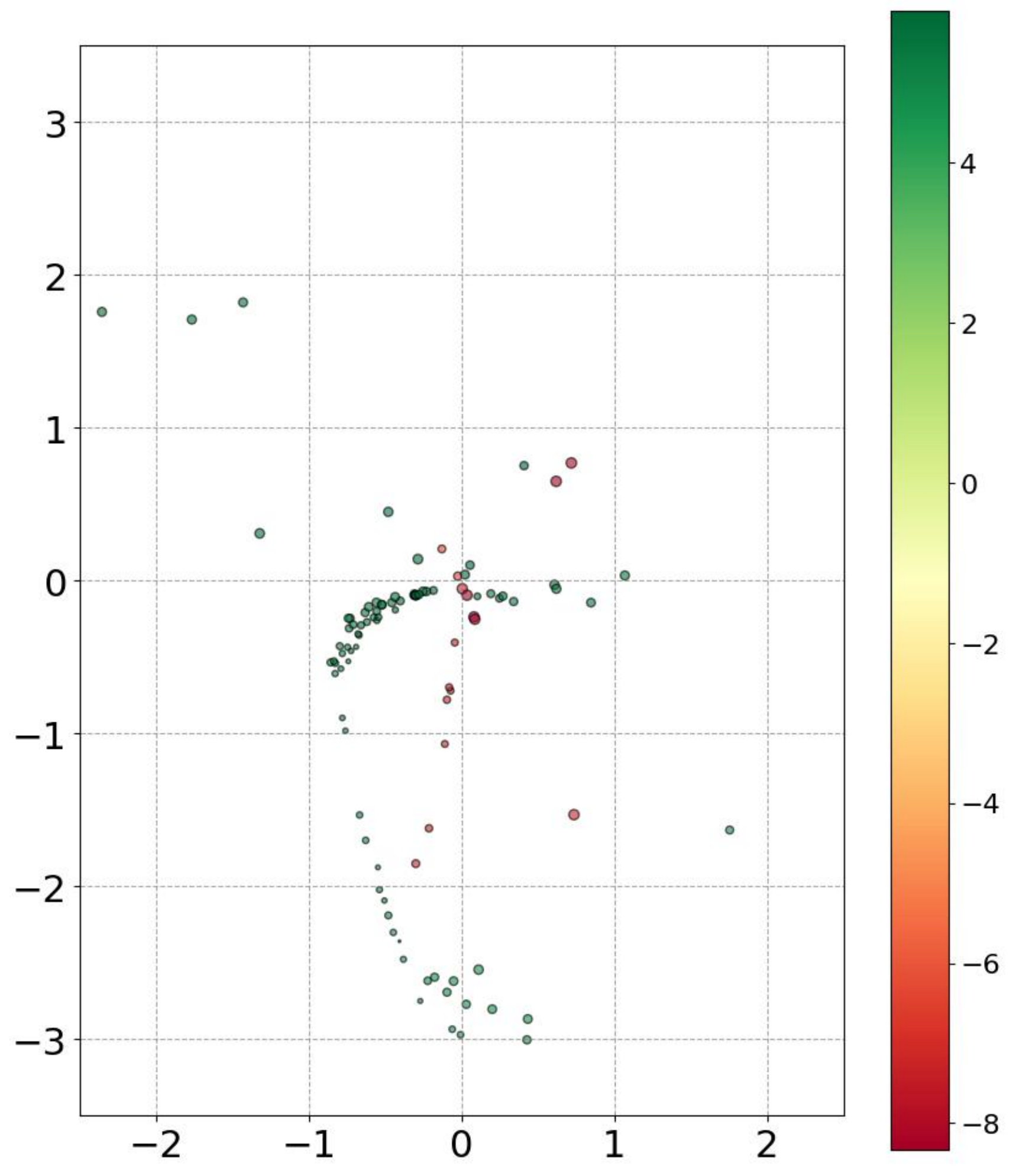}
} &
\bmvaHangBox{
    \includegraphics[width=0.225\columnwidth]{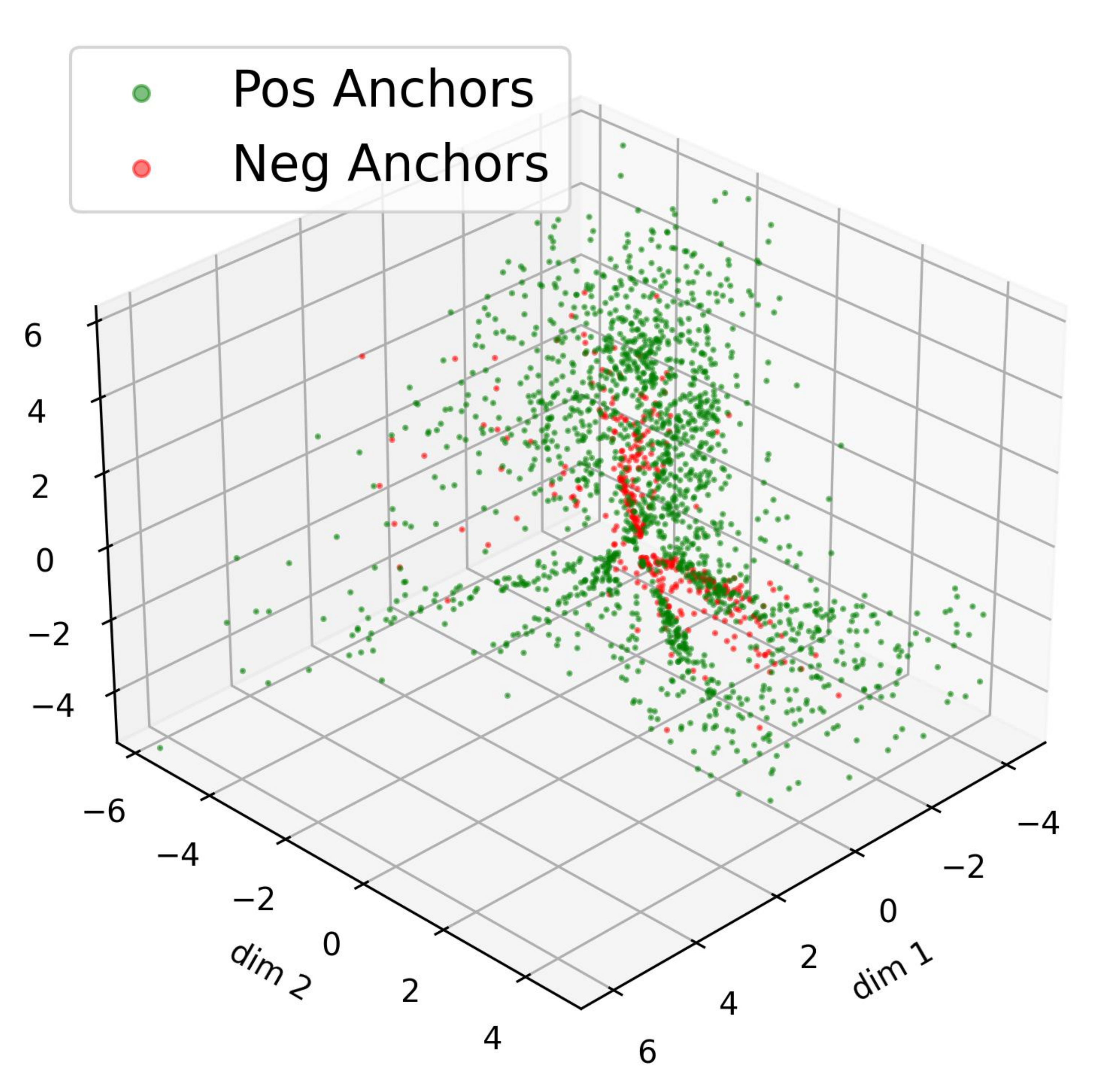}
} &
\bmvaHangBox{
    \includegraphics[width=0.225\columnwidth]{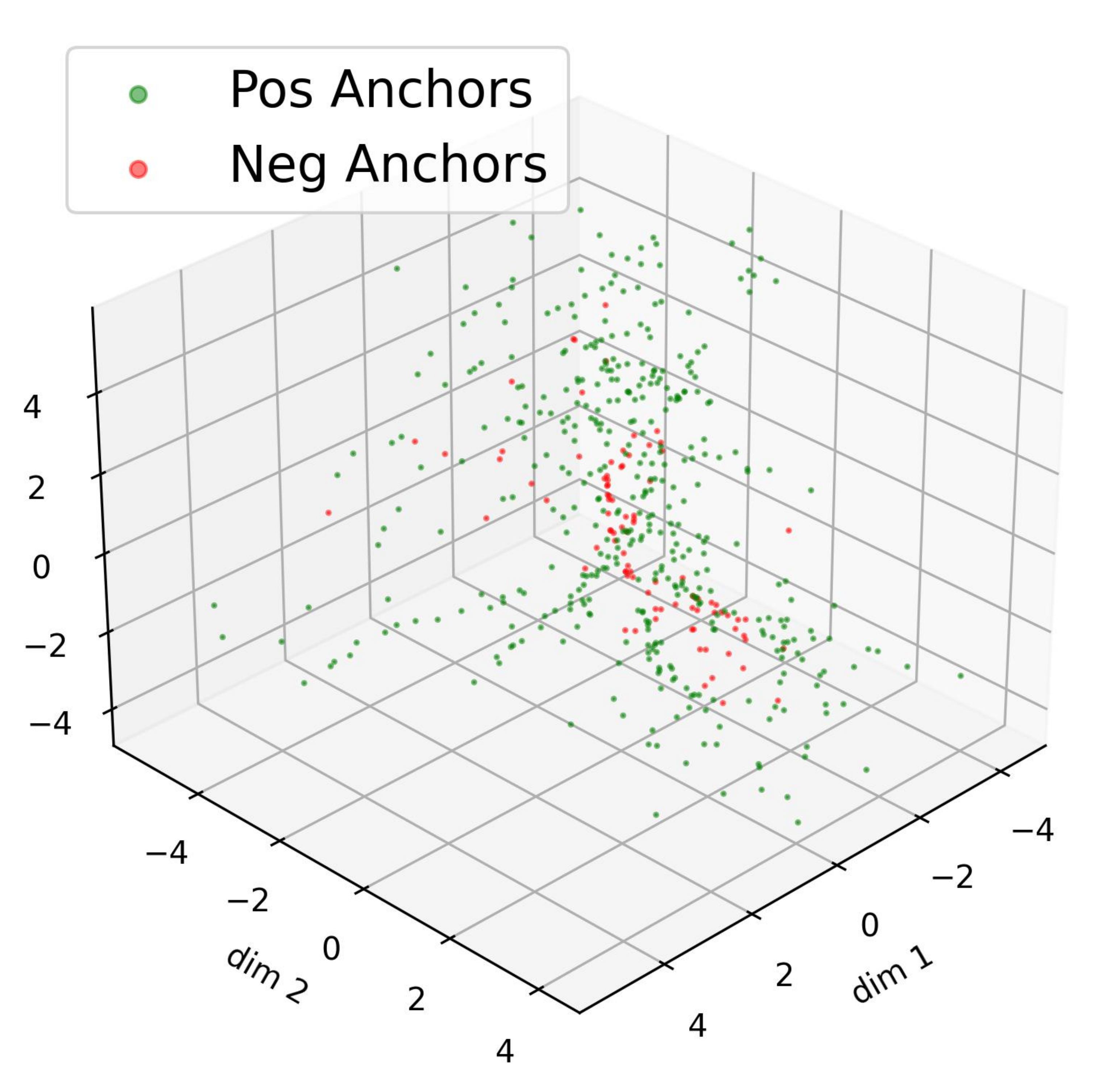}
}
\\
(a) 500 anchors & (b) 100 anchors & (c) 2000 anchors & (d) 500 anchors
\end{tabular}
\caption{
Comparison of learnt anchors from ResNet18 on the Waterbirds-WILDS with 500 vs.\ 100 anchors (a and b), and on CIFAR-10 remapped to a three-class problem with 2000 vs.\ 500 anchors (c and d). The color represents the peak values, while the point size reflects the extent of influence. Note that in our experiments, the influence decays according to the cosine distance rather than the Euclidean distance, as might be intuitive from the image.
}
\label{fig:anchors_location}
\end{figure}

This experiment illustrates the learnt anchors on two settings: the two-class case on Waterbirds-WILDS and the synthetic three-class case, which is remapped to three classes on CIFAR-10, as shown in Figure~\ref{fig:anchors_location} (a, b) and (c, d), respectively. The anchor positions, their peak values (indicated by color), and variance values (indicated by point size) are shown.

A clear aggregation pattern emerges, where positive and negative anchors (anchors with positive or negative peak values) cluster in distinct regions of logit space. This pattern suggests that logits with similar predictive performance tend to group together. Moreover, the consistent distribution of anchors across different anchor counts indicates that they consistently align with the same underlying distribution in the training data, further verifying the validity of our method.

\section{Conclusion}

In this work, ALSA, a framework designed to better understand the distribution and probability of correct predictions within the logit space, was proposed. The approach is motivated by the observation that models often exhibit imbalanced prediction performance across different classes—an issue overlooked by existing methods. By analyzing the properties of logits distributions, it is demonstrated that this information can be captured using only a limited number of anchors. Extensive experiments across diverse datasets and domains, including vision, language, and graph tasks, validate the effectiveness of ALSA in accurately estimating model performance. The results show that ALSA not only provide a more robust method for accuracy estimation but also offer deeper insights into model behavior, making them highly applicable for real-world machine learning deployments.

\section*{Acknowledgement}
This research has been partially supported by Australian Research Council Discovery Projects (DP230101196, DE250100919, CE200100025, and FT230100426).

\bibliography{egbib}

\clearpage
\appendix

\section*{Appendix}

\section{Related Work}
\label{appendix:related_work}

This section reviews methods of detecting out-of-distribution data and evaluating model generalization under distribution shifts.

\subsection{Out-of-Distribution Detection}

Out-of-distribution (OOD) detection identifies inputs outside the training distribution, improving model robustness in open-world scenarios. Confidence-based methods, such as using softmax scores, are widely applied~\cite{hendrycks2016baseline, geifman2017selective}. To mitigate overconfidence, temperature scaling~\cite{liang2017enhancing} adjusts logits for improved ID-OOD separation. Other methods analyze softmax discrepancies~\cite{jiang2018trust} or combine embeddings with density estimators~\cite{zhang2020hybrid}. Energy-based models (EBMs) provide an alternative by assigning energy scores so that ID data exhibits lower energy than OOD samples~\cite{du2019implicit, zhai2016deep, grathwohl2019your, elflein2021out,gold,topo}.

\subsection{Evaluating Out-of-Distribution Generalization}

Due to performance degradation under distribution shifts, evaluating generalization is crucial for reliable deployment~\cite{recht2019imagenet,koh2021wilds,geirhos2018imagenet}. Quantitative scores are commonly used for assessment. For instance, Confidence and Dispersity~\cite{deng2023confidence} employs the nuclear norm to measure class dispersion, while studies have shown a strong correlation between model agreement and generalization~\cite{nakkiran2020distributional, jiang2021assessing}. Accuracy-on-the-line~\cite{miller2021accuracy} links in-distribution (ID) performance to OOD generalization, and ProjNorm~\cite{yu2022predicting} predicts OOD error by analyzing model parameter sensitivity to shifts. Agreement-on-the-line~\cite{baek2022agreement} further estimates performance based on the agreement rate between ID and OOD predictions.

Other approaches estimate accuracy on unlabeled OOD datasets by using softmax confidence scores, such as ATC~\cite{garg2022ATC}, DoC~\cite{guillory2021predicting}, and AC~\cite{garg2022ATC}. AutoEval~\cite{deng2021labels} quantifies dataset differences via Frechet Distance, while a multi-task network~\cite{deng2021does} leverages rotational robustness for OOD accuracy estimation. For graphs, GNNEvaluator~\cite{zheng2024gnnevaluator} uses augmentation and discrepancy metrics to assess distributional shifts and predict OOD performance.

Unlike existing approaches that rely on softmax-based metrics, which compress critical information, or on data similarity measures, which are computationally intensive and tailored to specific domains, ALSA directly leverages anchors in logit space. By preserving richer predictive signals, ALSA provides a unified and robust framework for accurately estimating model performance under diverse distribution shifts encountered in real-world scenarios.

\section{Estimation of Band Width}
\label{appendix:proof_sec3}

\textbf{Proposition}. Denote the linear transformation as $\mathbf{z} = W\mathbf{h} + \mathbf{b}$, where $W \in \mathbb{R}^{c \times m}$. Assuming that $W_{ij} \sim N(0, \sigma_W^2)$ and that each element in the initial $\mathbf{h}$ is independent with mean $\mu_h = 0$ and variance $\sigma_h^2$, and that the bias $\mathbf{b}$ is initialized to zeros, applying Xavier initialization to the linear transformation and assuming $\mathbf{h}$ satisfies the variance condition of Xavier initialization, it would yield a band width less than $4z \sqrt{\frac{1}{m+c}}$, where \( z \) is the z-score of a Gaussian distribution

Let $W = [\mathbf{w}_1, \mathbf{w}_2, \ldots, \mathbf{w}_m]$. Since $W_{ij} \sim N(0, \sigma_W^2)$, and each element in the initial $\mathbf{h}$ is independent with mean $\mu_h = 0$ and variance $\sigma_h^2$, the bias $\mathbf{b}$ is initialized to all zeros. The following holds:
\begin{equation}
\mathbf{z} = W\mathbf{h} + \mathbf{b} = \sum_{i=1}^{m} \mathbf{w}_i h_i
\end{equation}
To compute the variance of the logits $\mathbf{z}$, the following expression is obtained:
\begin{equation}
\begin{aligned}
\mathrm{var}(\mathbf{z}) &= \mathrm{var}\left(\sum_{i=1}^{m} \mathbf{w}_i h_i\right) = \sum_{i=1}^{m} \mathrm{var}(\mathbf{w}_i h_i) \\
 &= \sum_{i=1}^{m} \big( [E(h_i)]^2 \mathrm{var}(\mathbf{w}_i) + [E(\mathbf{w}_i)]^2 \mathrm{var}(h_i)  \\    
 &\quad + \mathrm{var}(\mathbf{w}_i)\mathrm{var}(h_i) \big)
\end{aligned}    
\end{equation}

Since $E(h_i) = 0$ and $E(\mathbf{w}_i) = 0$, the equation simplifies to
\begin{align}
\mathrm{var}(\mathbf{z}) &= \sum_{i=1}^{m} \mathrm{var}(\mathbf{w}_i)\mathrm{var}(h_i) = m \sigma_W^2 \sigma_h^2
\end{align}
Similarly, for each $z_i$, the following holds:
\begin{align}
\mathrm{var}(z_i) &= \sum_{j=1}^{m} \mathrm{var}(W_{ij} h_j) = \sum_{j=1}^{m} \mathrm{var}(W_{ij}) \mathrm{var}(h_j) = m \sigma_W^2 \sigma_h^2
\end{align}

The variance along the direction of the all-ones vector follows the same rules. Let the length of the projection of the logit vector $\mathbf{z}_i$ onto the all-ones vector be denoted as $l$; the following is obtained:
\begin{align}
l = \frac{1}{\sqrt{c}} \mathbf{1}^\top \mathbf{z}_i = \frac{1}{\sqrt{c}} \sum_{i=1}^c z_i
\end{align}
Then,
\begin{align}
\mathrm{var}(l) &= \frac{1}{c} c \, \mathrm{var}(z_i) = m \sigma_W^2 \sigma_h^2
\end{align}
In practice, to ensure the stability of the gradient, initialization techniques, such as Xavier \cite{glorot2010understanding}, are used. If Xavier initialization is applied with $\sigma_W^2 = \frac{2}{m + c}$, and $\mathbf{h}$ also follows the variance of Xavier initialization with $\sigma_h^2 < \frac{2}{m}$.

Therefore, the following holds:
\begin{align}
\mathrm{var}(l) &= m \sigma_W^2 \sigma_h^2 \\
&< \frac{4}{m + c}
\end{align}
The width of the band is less than $2z\sqrt{\frac{4}{m + c}} = 4z\sqrt{\frac{1}{m + c}}$, where $z$ is the z-score corresponding to any prescribed confidence interval of a Gaussian distribution.

For a linear classifier taking a 128-dimensional embedding as input and solving a 10-class classification problem, it is expected to have 99.9\% confidence that the width of the band near the hyperplane is less than
\begin{equation}
    2 \times 3.291 \times \sqrt{\frac{4}{128 + 10}} = 1.121,
\end{equation}

where 3.291 is the z-score corresponding to a 99.9\% confidence interval.

\section{Logits Distribution in a Three-Class Scenario}

\label{appendix:logits_distribution_3d}

\begin{figure*}[h]
\centering
\begin{tabular}{cccc}
\bmvaHangBox{
    \includegraphics[width=0.18\textwidth]{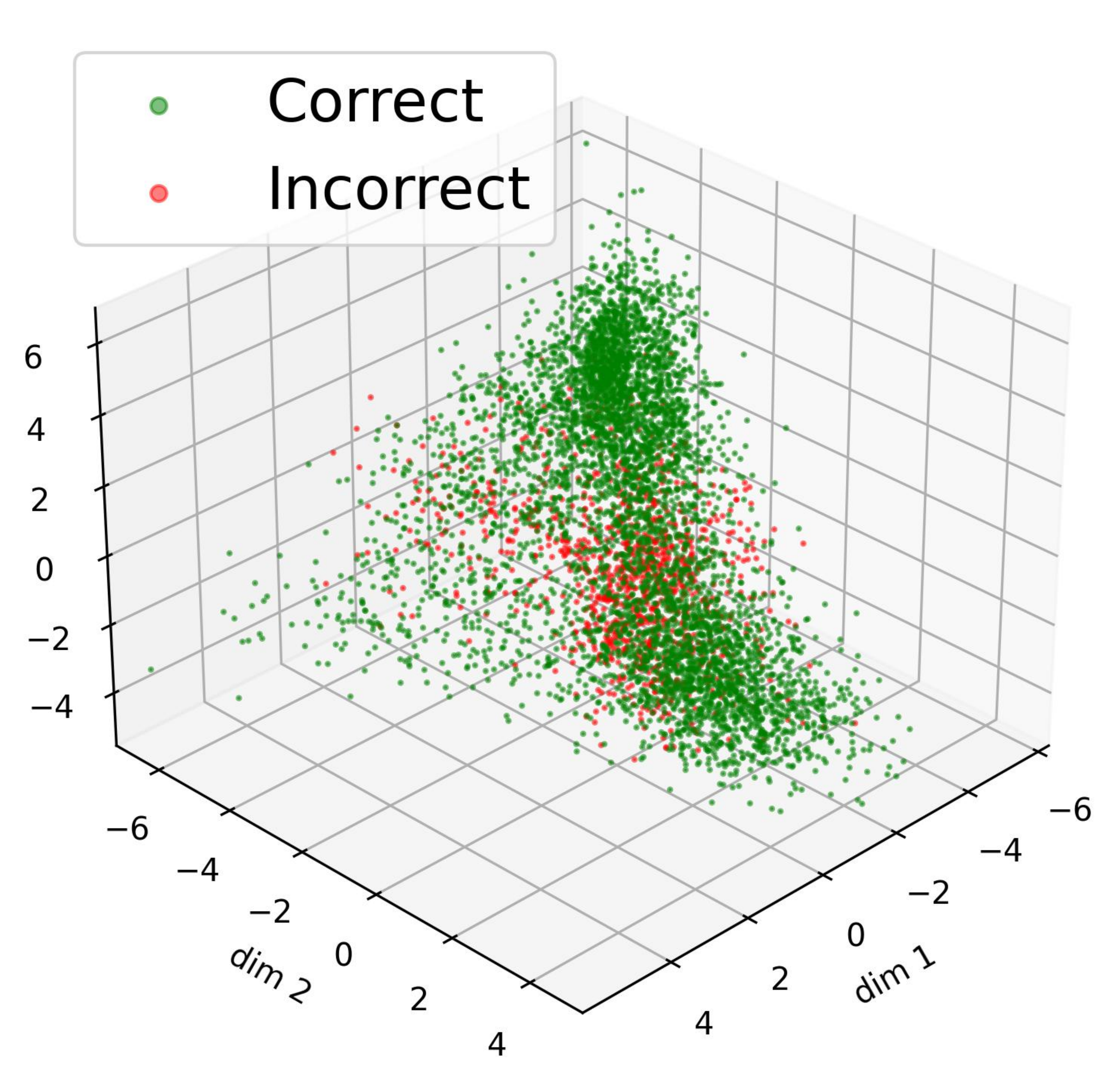}
} &
\bmvaHangBox{
    \includegraphics[width=0.18\textwidth]{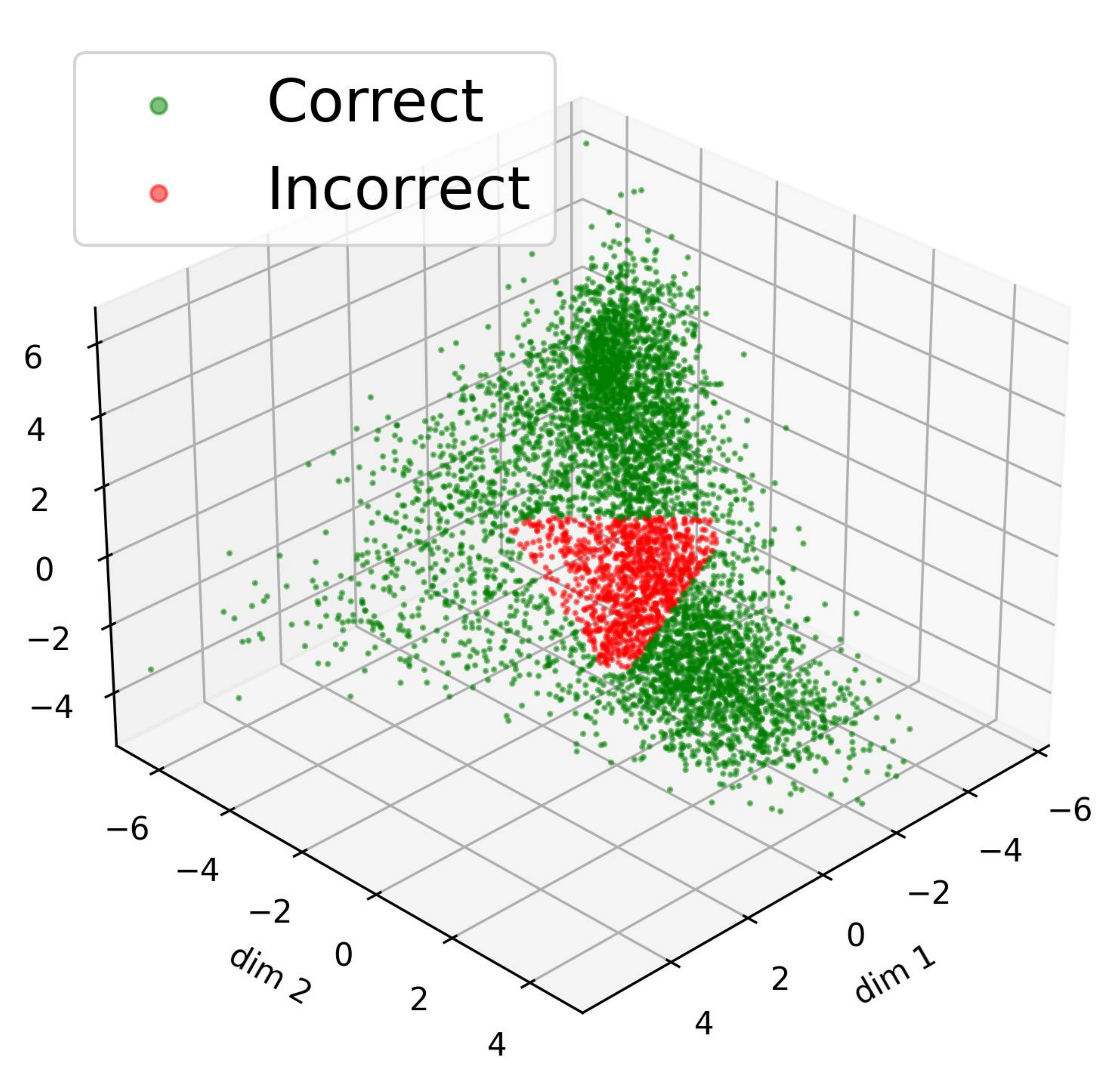}
} &
\bmvaHangBox{
    \includegraphics[width=0.22\textwidth]{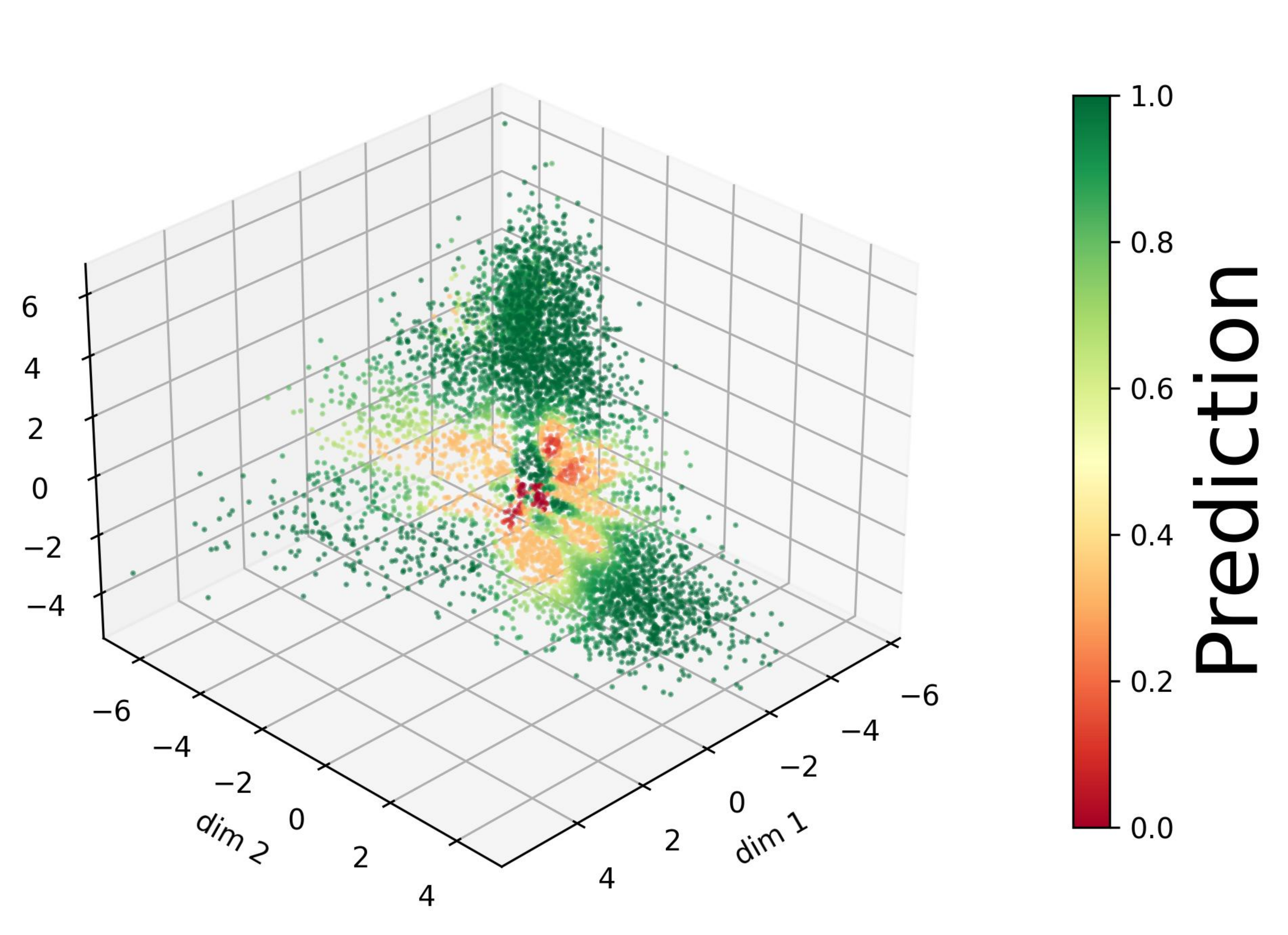}
} &
\bmvaHangBox{
    \includegraphics[width=0.22\textwidth]{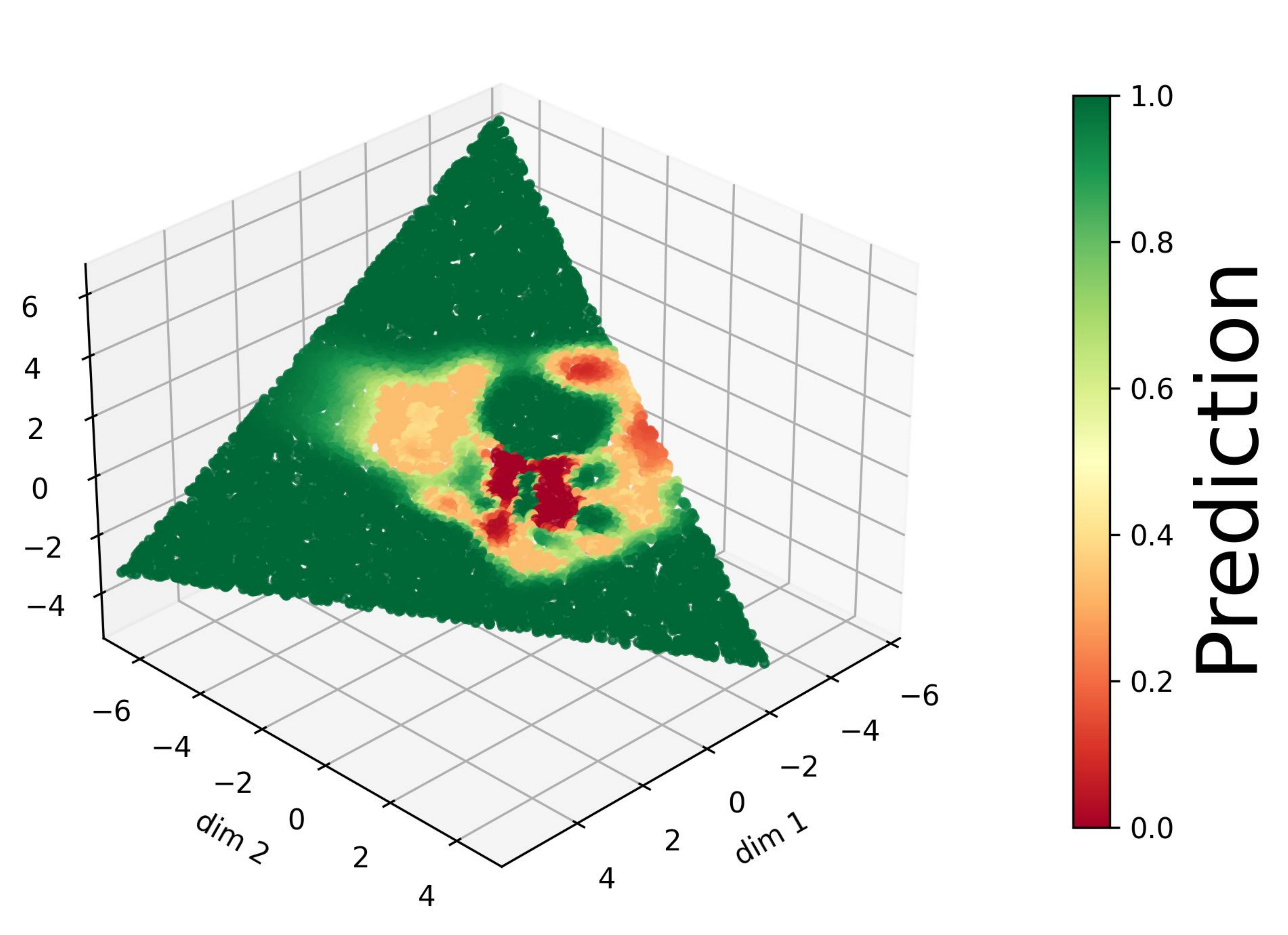}
} \\
(a) True pred.  &
(b) ATC pred. &
(c) ALSA pred. scatter  &
(d) ALSA pred. surface 
\end{tabular}
\caption{Visualization of the logits distribution and prediction estimations on the validation set for a three-class classification problem using ATC and ALSA. The logits are generated by remapping the CIFAR-10 dataset into a three-class classification task. (a) Each point represents a logit, with color indicating prediction correctness: green for correct predictions and red for incorrect predictions. (b) Predictions made by ATC: red points represent samples identified as incorrect, while green points represent those marked as correct. (c) The estimated probability that a logit corresponds to a correct prediction at various positions, as inferred by ALSA from the logits shown in Figure (a). (d) ALSA prediction surface plot on the best-fit plane of logits on validation set.}
\label{fig:3d_vis}
\end{figure*}

The logits distribution for a three-class classification problem is visualized in Figure~\ref{fig:3d_vis} (a) and (b). 

Similar to the binary case, where logits align closely along a linear structure, in this scenario, they are primarily distributed near a plane in 3D space. This suggests that the logits distribution naturally forms a lower-dimensional region of the logit space. As in the binary case, the likelihood of correct predictions varies depending on logit positions. The incorrect predictions in Figure~\ref{fig:3d_vis}(a) show the skewness between classes, further highlighting the non-uniformity of model performance across classes.

These observations reinforce the hypothesis that logits encode structured information about prediction confidence, which is not fully captured by studying softmax output. The ALSA framework is introduced to analyze this structure and leverage logit positioning for improved predictive reliability estimation.

\section{Comparison of Initialization Methods}
\label{appendix:init_method}

\begin{figure}[h]
\centering
\begin{tabular}{cc}
\bmvaHangBox{
    \includegraphics[width=0.45\columnwidth]{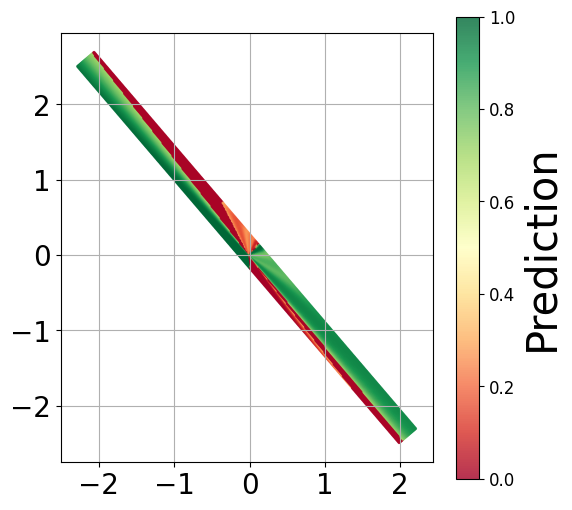}
} &
\bmvaHangBox{
    \includegraphics[width=0.45\columnwidth]{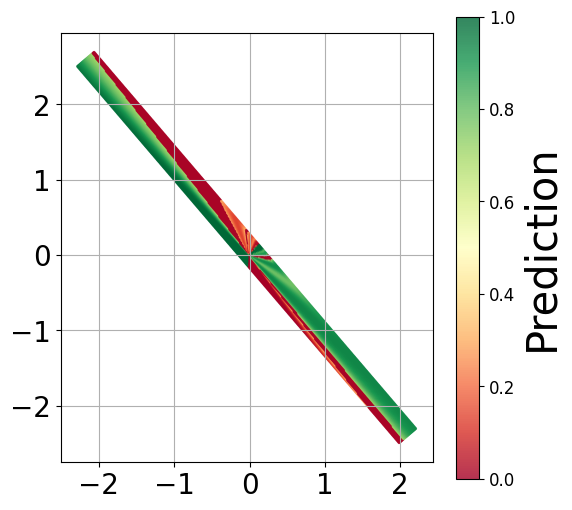}
} \\
(a) Sampling & (b) Random Initialization
\end{tabular}
\caption{Comparison of sampling and random initialization.}
\label{fig:init_method}
\end{figure}

ALSA is not sensitive to initialization because its objective is to learn anchors that align with the underlying logit distribution. Given the same training data, different initialization methods should ideally converge to a similar distribution. Furthermore, anchor positions are continuously refined during training, ultimately converging to a consistent target distribution regardless of the initialization strategy.

In Figure~\ref{fig:init_method}, we compare the ALSA prediction surface plots in the region where logits are distributed under two different initialization policies: sampling anchor positions from the validation set and randomly initializing them from a normal distribution. As shown in Figure~\ref{fig:init_method}(a) and Figure~\ref{fig:init_method}(b), both methods yield nearly identical learnt probability distributions in logit space, demonstrating the robustness of ALSA to initialization choices.

\section{Rationale of Unified Threshold t for Rectification}
\label{appendix:anchors_t}

To rectify the prediction in regions of the logit space where ALSA cannot exert reliable influence, the estimation is adjusted. The intuition behind this correction is straightforward: if no single anchor provides sufficient influence in a particular region, the total influence in that region approaches zero, yielding a predicted accuracy close to 0.5. In this case, it is assumed that the model behaves like a random guesser in the region, and thus, the predicted accuracy is adjusted to $\frac{1}{c}$, where $c$ is the number of classes.

Insufficient influence is formally defined as follows:

\textit{For any logits $\mathbf{z}_i=f(\mathbf{x}_i)$, where $\mathbf{x}_i$ is an input sample, and the set of anchors $\mathcal{A} = \{ (\mathbf{a}_i, p_i, v_i) \}_{i=1}^k$, if for all $j \in \{1, 2, \dots, k\}$, we have $|\mathrm{Infl}(\mathbf{z}_i, (\mathbf{a}_j, p_j, v_j))| < t$, then the anchor set $\mathcal{A}$ cannot provide sufficient influence on logit $\mathbf{z}_i$.}

For any single anchor, its influence decays as the distance increases. The Area Under the Curve (AUC) is tracked to assess how the influence diminishes. A hyperparameter $\alpha \in (0, 1)$ is introduced to control the confidence interval in the decayed region.

Formally, for the Gaussian-like influence function, the distance between any point $\mathbf{z}_i$ in logit space is denoted as $w = \mathrm{dist}(\mathbf{z}_i, \mathbf{a}_j)$. When $\mu$ satisfies
\begin{equation}
    \int_0^{\mu} p_j \exp(-v_j^2 w^2) dw = \alpha \int_0^\infty p_j \exp(-v_j^2 w^2) dw,
\end{equation}
$t = p_j \exp(-v_j^2 \mu^2)$ is set. Here, $t$ represents the influence value when the AUC has decayed by $\alpha$ of its total value over $(0, \infty)$.

The solution for $t = p_j \exp(-v_j^2 \mu^2)$ is equivalent to $t = p_j \exp(-\eta^2)$ when
\begin{align}
    \int_0^{\eta} p_j \exp(-r^2) \frac{dr}{v_j} = \alpha \int_0^\infty p_j \exp(-r^2) \frac{dr}{v_j},
\end{align}
where $r = v_j w$.

According to~\cite{andrews1998special}, this equation has no simple closed-form solution but can be expressed using the error function $\mathrm{erf}(\cdot)$ as
\begin{equation}
    \text{LHS} = \frac{p_j}{v_j} \int_0^{\eta} \exp(-r^2) dr = \frac{p_j}{v_j} \frac{\sqrt{\pi}}{2} \mathrm{erf}(\eta),
\end{equation}
and
\begin{equation*}
    \text{RHS} = \frac{p_j}{v_j} \int_0^\infty \exp(-r^2) dr = \frac{p_j}{v_j} \alpha \frac{\sqrt{\pi}}{2}.
\end{equation*}

Therefore, $\eta = \mathrm{erf}^{-1}(\alpha)$ is obtained and consequently
\begin{equation}
    t = p_j \exp\left(-\left( \mathrm{erf}^{-1}(\alpha) \right)^2\right).
\end{equation}

Similarly, for the exponential influence function, when $\mu$ satisfies
\begin{equation}
    \int_0^{\mu} p_j \exp(-v_j^2 w) dw = \alpha \int_0^\infty p_j \exp(-v_j^2 w) dw,
\end{equation}
$t$ is set as $t=p_j\exp(-v^2_j\mu)$. The solution for $t$ is given by  $t=p(1-\alpha)$. 

The reliable maximum peak value is set to 6, as this value corresponds to a predicted accuracy of 99.75\% after applying the sigmoid function, which is sufficiently high for a confident prediction. Anchors with smaller peak values may still fine-tune probabilities in smaller regions, but they cannot exert sufficient influence on their own. Hence, using a unified peak value is justified. A value of 6 is chosen for all test cases, and the cut-off value \(t\), applicable to all anchors, is given by  $t = 6 \exp(-(\mathrm{erf}^{-1}(\alpha))^2)$ for the Gaussian-like influence function and  $t = 6(1-\alpha)$ for the exponential influence function.

\section{Dataset Details}

\label{appendix:dataset_details}

In the comprehensive evaluation, benchmark datasets across multiple types of distributions commonly encountered in real-world scenarios, including vision, language, and graph datasets, were considered. Unlike vision and language datasets, graphs represent non-Euclidean data, and the tests were extended to include this type of data. For datasets with an evenly distributed label marginal distribution, sampling was applied to simulate imbalanced class distributions to better reflect real-world scenarios.

\textbf{Natural shift.} Non-simulated shifts were first considered, where distributional changes arise from variations in the data collection process. Three vision datasets were included to evaluate this type of shift. CIFAR-10.1~\cite{recht2018cifar101} and CIFAR-10.2~\cite{lu2020harder} are derived from the same source as CIFAR-10~\cite{krizhevsky2009learning} and were assembled via a similar process. ImageNetV2~\cite{recht2019imagenet} was sampled a decade after the original ImageNet~\cite{deng2009imagenet} dataset.

\textbf{Subpopulation shift.} Subpopulation shift refers to the distributional changes between training and test data, where specific subgroups or categories within the data change in proportion or characteristics, leading to reduced model performance on underrepresented or shifted subgroups. In the experiments, two WILDS datasets~\cite{wilds2021} were used as benchmarks. The Waterbirds-WILDS~\cite{sagawa2019distributionally} dataset was considered for vision tasks, and Amazon-WILDS was used for language tasks. Amazon-WILDS provides an OOD validation dataset and an OOD test set, both of which were used as OOD datasets for training, referred to as \texttt{Subpop1} and \texttt{Subpop2}, respectively.

\textbf{Synthetic Shift.} Synthetic shifts are often caused by deliberate corruption. For vision tasks, common corruptions include changes in brightness, blurriness, and saturation, which frequently occur in real-world scenarios. In the experiments, CIFAR-10C~\cite{hendrycks2019benchmarking} and CIFAR-100C~\cite{hendrycks2019benchmarking}, which contain 19 different corruptions, each with 5 levels of severity, were used. The MNIST-M dataset~\cite{zhao2020review}, a modified version of the MNIST dataset where digits are blended with random colored background patches from natural images, introducing more complex visual patterns, was also evaluated.

\textbf{Domain shift.} Domain shift refers to the change in data distribution between the source domain and the target domain, which can lead to model performance degradation, as the model is trained on data that does not fully represent the new domain. In the experiments, Office-31~\cite{koniusz2017domain}, which contains data sampled from three different domains with the same classes of definition, was used, posing a challenge to the model's generalization and adaptation ability across different domains.

\textbf{Temporal shift.} Temporal shift refers to changes in data distribution over time, where the characteristics of the data at training time differ from those at inference or testing time. This can lead to degraded model performance as the model may not generalize well to time-evolved data. The focus was on the ogbn-arxiv~\cite{hu2020ogb} dataset, a graph dataset for paper citation networks, containing data from 1971 to 2020.

\section{Baseline Details}
\label{appendix:baseline_details}

The method, ALSA, is compared with an array of baselines.

\textit{Average Confidence (AC)}~\cite{garg2022ATC} estimates the target set error by taking the average of maximum softmax confidence:
\begin{equation}
    \hat{Acc}(f, D^T) = \mathbb{E}_{x \sim D^T}\left[\max_{j \in \mathcal{Y}} f_j(x)\right].
\end{equation}
\textit{Difference of Confidence (DoC)}~\cite{guillory2021predicting} estimates the error through the difference between the confidence of source data and the confidence of target data:
\begin{equation}
\begin{aligned}
\hat{Acc}(f, D^T) &= \mathbb{E}_{x \sim \mathcal{D}_S} \left[ \mathbb{I} \left[ \arg \max_{j \in \mathcal{Y}} f_j(x) \neq y \right] \right] \\
\quad &+ \mathbb{E}_{x \sim \mathcal{D}_T} \left[ \max_{j \in \mathcal{Y}} f_j(x) \right] - \mathbb{E}_{x \sim \mathcal{D}_S} \left[ \max_{j \in \mathcal{Y}} f_j(x) \right].
\end{aligned}
\end{equation}

\textit{Importance re-weighting (IM)} estimates the error of the classifier with importance re-weighting of 0-1 error in the pushforward space of the classifier. This corresponds to MANDOLIN using one slice based on the underlying classifier confidence~\cite{chen2021mandoline}.

\textit{Average Thresholded Confidence (ATC)}~\cite{garg2022ATC} identifies a threshold $t$ such that the fraction of source data points that have scores below the threshold matches the source error on in-distribution validation data. Target error is estimated as the expected number of target data points that fall below the identified threshold:
\begin{equation}
    \hat{Acc}(f, D^T) =\mathbb{E}_{x \sim D_T} \left[\mathbb{I} \left[ -H(\vec{f}(x)) < t \right] \right],
\end{equation}
where $H(\cdot)$ is the negative entropy function mapping the softmax vector to a scalar.

\textit{Confidence Optimal Transport (COT)}~\cite{lu2023characterizing} addresses out-of-distribution (OOD) error estimation by calculating the optimal transport cost between the predicted and true label distributions. This cost reflects the discrepancy necessary to transform one distribution into the other, effectively measuring the error in model predictions on OOD data.

\textit{GNNEvaluator}~\cite{zheng2024gnnevaluator} evaluates the performance of graph neural networks (GNNs) by first constructing a simulated dataset of graphs (DiscGraph set) to capture potential distribution discrepancies. It then trains a model (GNNEvaluator) to estimate the node classification accuracy on these graphs, effectively predicting GNN performance under real-world conditions without requiring labels.

\section{Details on the Experimental Setup}
\label{appendix:exp_setup}

Experiments were run on RTX 2080TI and A100 GPUs. PyTorch~\cite{paszke2019pytorch} was used for all experiments.

\subsection{Architectures and Evaluations} 
For vision tasks, ResNet18~\cite{he2016deep} is trained on CIFAR-10, CIFAR-100, Office-31, and Waterbirds-WILDS. Pretrained AlexNet, MobileNet-v2, and EfficientNet-B0 from PyTorch~\cite{pytorch2024models} are used on ImageNet. For language tasks, DistilBERT-base-uncased~\cite{sanh2019distilbert} is fine-tuned on Amazon-WILDS. For graph tasks, GCN~\cite{kipf2016semi}, GIN~\cite{xu2018powerful}, GAT~\cite{velickovic2017graph}, and SAGE~\cite{hamilton2017inductive} are trained on ogbn-arxiv to compare estimation methods across architectures. Temperature Scaling (TS)~\cite{guo2017calibration} is applied to in-distribution validation sets for all methods except ALSA, which does not require TS. TS adjusts neural network output probabilities to better reflect correctness likelihood and improves softmax-based accuracy estimation~\cite{garg2022ATC}. Performance is assessed by the Mean Absolute Error (MAE) between true and estimated accuracy on out-of-distribution (OOD) datasets (the lower the better).

\subsection{Training Details}
\textbf{CIFAR10 and CIFAR100}: Both the CIFAR10 and CIFAR100 datasets were subsampled to create imbalanced versions for training a ResNet18 model from scratch. Specifically, the ratio of samples between Class 0 and Class 9 was set to 1:3, and this imbalance increased linearly across the classes from Class 0 to Class 9. This method was used to introduce an uneven distribution of samples across the classes. Weight decay was set to $5\times10^{-4}$ and the batch size to 200, with early stopping after 10 epochs. The initial learning rate was set to 0.1 and multiplied by 0.1 every 100 epochs.

\textbf{ImageNet}: Due to limited computational resources, three pretrained models—AlexNet, MobileNetV2, and EfficientNetB0—provided by PyTorch were directly utilized. The official ImageNet validation set was used for evaluation.

\textbf{Amazon-WILDS}: The \texttt{id\_val} set was used for validation, and a DistilBERT-base-uncased model~\cite{sanh2019distilbert} was trained. Training was performed using the AdamW optimizer~\cite{loshchilov2017decoupled} with a weight decay of $10^{-2}$, a learning rate of $10^{-5}$, and a batch size of 8, over 3 epochs. The maximum sequence length was limited to 512 tokens.

\textbf{Waterbirds-WILDS}: The \texttt{val} set was used for validation, and a ResNet18 model was trained with a batch size of 200, applying early stopping after 5 epochs. The Adam optimizer was used with a learning rate of $10^{-3}$.

\textbf{Office-31}: A ResNet18 was trained on data sampled from the Amazon domain, using 80\% of the data for training and 20\% for validation, with a batch size of 64. The model was optimized using Adam with a fixed learning rate of $10^{-3}$, and training was conducted for 300 epochs with early stopping after 5 epochs if the validation accuracy did not improve.

\textbf{MNIST}: A fully connected network with three convolutional layers was trained, using the Adam optimizer with a learning rate of $10^{-3}$ and early stopping after 5 epochs. To introduce class imbalance, the MNIST dataset was subsampled, setting the ratio of samples between Class 0 and Class 9 to 1:100, with the ratio increasing linearly across the remaining classes.

\textbf{ogbn-arxiv}: Four different models were trained: GCN, GAT, GIN, and SAGE, using model architectures provided by PyTorch. Data before 2011 was used for training, validation was performed on data from 2012 to 2014, and testing was conducted on data from 2015 to 2020. For all models, the Adam optimizer was used with a fixed learning rate of $10^{-2}$ and a weight decay of $5 \times 10^{-4}$, along with early stopping if validation accuracy did not improve after 20 epochs.

\section{Detailed Experiment Results}
\label{appendix:exp_details}

\subsection{Full Table of Vision and Language Datasets Results}

Comprehensive experiments were conducted on vision and language datasets. In Table~\ref{tab:vision_language_mae_std}, the Mean Absolute Estimation Error (MAE) between the estimated and true accuracy across different datasets in the setup is reported. The standard deviation (std) for each experiment is also included, with the results presented in parentheses.

\begin{table}[htbp]
\centering
\small
\caption{Mean Absolute Estimation Error (MAE) between the estimated and true accuracy across different datasets in our setup. The results are averaged over three random seeds. Standard deviations are provided below each value.}
\label{tab:vision_language_mae_std}
\resizebox{\columnwidth}{!}{%
\begin{tabular}{llccccccc}
\toprule
\textbf{Dataset} & \textbf{Variant} & \textbf{AC (\%)} & \textbf{DoC (\%)} & \textbf{IM (\%)} & \textbf{ATC (\%)} & \textbf{COT (\%)} & \textbf{ALSA-E (\%)} & \textbf{ALSA-G (\%)} \\
\midrule
CIFAR10   & CIFAR-10.1   & 4.82  & 4.69  & 5.70  & 1.27  & 3.02  & 2.83 & \textbf{1.22}  \\
          &             & (0.95)& (0.90)& (0.75)& (0.78)& (0.89)& (1.39) & (1.97)\\
          \cmidrule(lr){2-9}
          & CIFAR-10.2   & 7.97  & 7.84  & 8.74  & 4.59  & 6.20  & \textbf{0.61} & 4.19\\
          &             & (0.21)& (0.14)& (0.04)& (0.66)& (0.21)& (0.54) & (0.66)\\
          \cmidrule(lr){2-9}
          & CIFAR-10C   & 6.57  & 6.44  & 7.60  & \textbf{2.28}  & 3.18 & 5.09 & 4.18  \\
          &             & (6.84)& (6.83)& (7.33)& (2.95)& (1.89)& (3.89) & (6.03)\\
\midrule
CIFAR100  & CIFAR-100C  & 9.46  & 7.80  & 9.02  & 3.85  & \textbf{2.41}  & 7.26 & 6.56  \\
          &             & (7.80)& (7.78)& (8.27)& (3.98)& (1.88) & (8.03) & (7.45)\\
\midrule
Amazon-WILD & Subpop1   & 3.70  & 2.16  & 2.14  & 5.90  & 3.39  & 0.99 & \textbf{0.97}  \\
            &           & (2.66)& (0.17)& (0.13)& (7.66)& (2.47) & (0.27) & (0.91)\\
            \cmidrule(lr){2-9}
            & Subpop2   & 4.19  & 2.65  & 2.65  & 5.82  & 4.18   & 1.43 & \textbf{1.17}  \\
            &           & (2.82)& (0.33)& (0.29)& (7.91)& (2.76) & (0.39) & (0.60)\\
\midrule
Waterbirds-WILDS & Subpop & 0.78  & 0.79  & 0.74  & 1.55  & 0.97  & \textbf{0.32} & 0.54  \\
                 &        & (0.93)& (0.36)& (0.43)& (0.44)& (1.11) & (0.18) & (0.30)\\
\midrule
ImageNet & ImageNetv2  & 3.63  & 2.68  & 2.85  & \textbf{0.75}  & 9.57  & 4.07 & 3.22  \\
         &             & (2.59)& (2.41)& (2.56)& (0.54)& (4.63)& (2.91) & (2.05)\\
\midrule
MNIST    & MNIST-M     & 13.27 & 13.44 & 16.29 & 13.63 & 4.96  & \textbf{2.82} & 3.09  \\
         &             & (3.54)& (3.51)& (2.00)& (3.45)& (2.52)& (3.73) & (1.73)\\
\midrule
Office-31    & domain     & 38.67 & 34.47 & 32.90 & 41.55 & 11.40  & 8.35 & \textbf{7.77}  \\
         &             & (15.29)& (8.35)& (7.37)& (33.25)& (6.44)& (7.47) & (3.32)\\
\bottomrule
\end{tabular}%
}
\end{table}

\subsection{Full Table of ogbn-arxiv Results}

Comprehensive experiments were conducted on the ogbn-arxiv dataset by evaluating four different models: GCN, GAT, GIN, and SAGE. In Table~\ref{tab:mae_gnn_std}, the Mean Absolute Estimation Error (MAE) on predictions for the years 2015 to 2020 is reported, along with the corresponding standard deviation to assess the consistency of the prediction performance.

\begin{table}[htbp]
\centering
\small 
\caption{Mean Absolute Estimation Error (MAE) and standard deviation (std) between the estimated and true accuracy across different datasets in our setup. The results are averaged over three random seeds. The best-performing method is highlighted.}
\label{tab:mae_gnn_std}
\resizebox{\columnwidth}{!}{%
\begin{tabular}{lcccccccc}
\toprule
\textbf{Model} & \textbf{AC (\%)} & \textbf{DoC (\%)} & \textbf{IM (\%)} & \textbf{GNNEval (\%)} & \textbf{ATC (\%)} & \textbf{COT (\%)} & \textbf{ALSA-E (\%)} & \textbf{ALSA-G (\%)} \\
\midrule
GCN     & 1.61  & 2.94  & 3.89  & 10.59 & 19.76  & 6.64 & 1.13 & \textbf{0.99}  \\
        & (1.09) & (1.54) & (1.43) & (8.81) & (5.76) & (2.18) & (1.18) &  (0.58) \\
\cmidrule(lr){2-9}
SAGE    & 1.31  & 1.72  & 2.07  & 10.42 & 25.31  & 5.85  & 2.21 &  \textbf{1.08}  \\
        & (0.56) & (0.67) & (0.94) & (8.42) & (3.21) & (2.72) & (1.55) &  (0.76) \\
\cmidrule(lr){2-9}
GAT     & 3.24  & 4.16  & 4.60  & 13.34 & 23.89  & 5.06  & 1.73 &  \textbf{1.01}  \\
        & (3.42) & (3.47) & (3.54) & (9.22) & (3.80) & (2.84) & (2.20) &  (0.86) \\
\cmidrule(lr){2-9}
GIN     & 30.09 & 5.86  & 7.17  & 9.54  & 32.28  & 30.79 & 2.18 &  \textbf{2.14}  \\
        & (2.72) & (2.70) & (1.63) & (6.49) & (4.28) & (2.36) & (1.23) &  (1.71) \\
\cmidrule(lr){2-9}
average & 9.06  & 3.67  & 4.43  & 10.97 & 25.31  & 12.09 & 1.81 &  \textbf{1.30}  \\
        & (1.95) & (2.10) & (1.89) & (8.24) & (4.26) & (10.1) & (1.54) &  (0.98) \\
\bottomrule
\end{tabular}%
}
\end{table}

\subsection{CIFAR-10C and CIFAR-100C Results Visualization}

\begin{figure*}
\centering
\begin{tabular}{cc}
\bmvaHangBox{
    \includegraphics[width=0.45\textwidth]{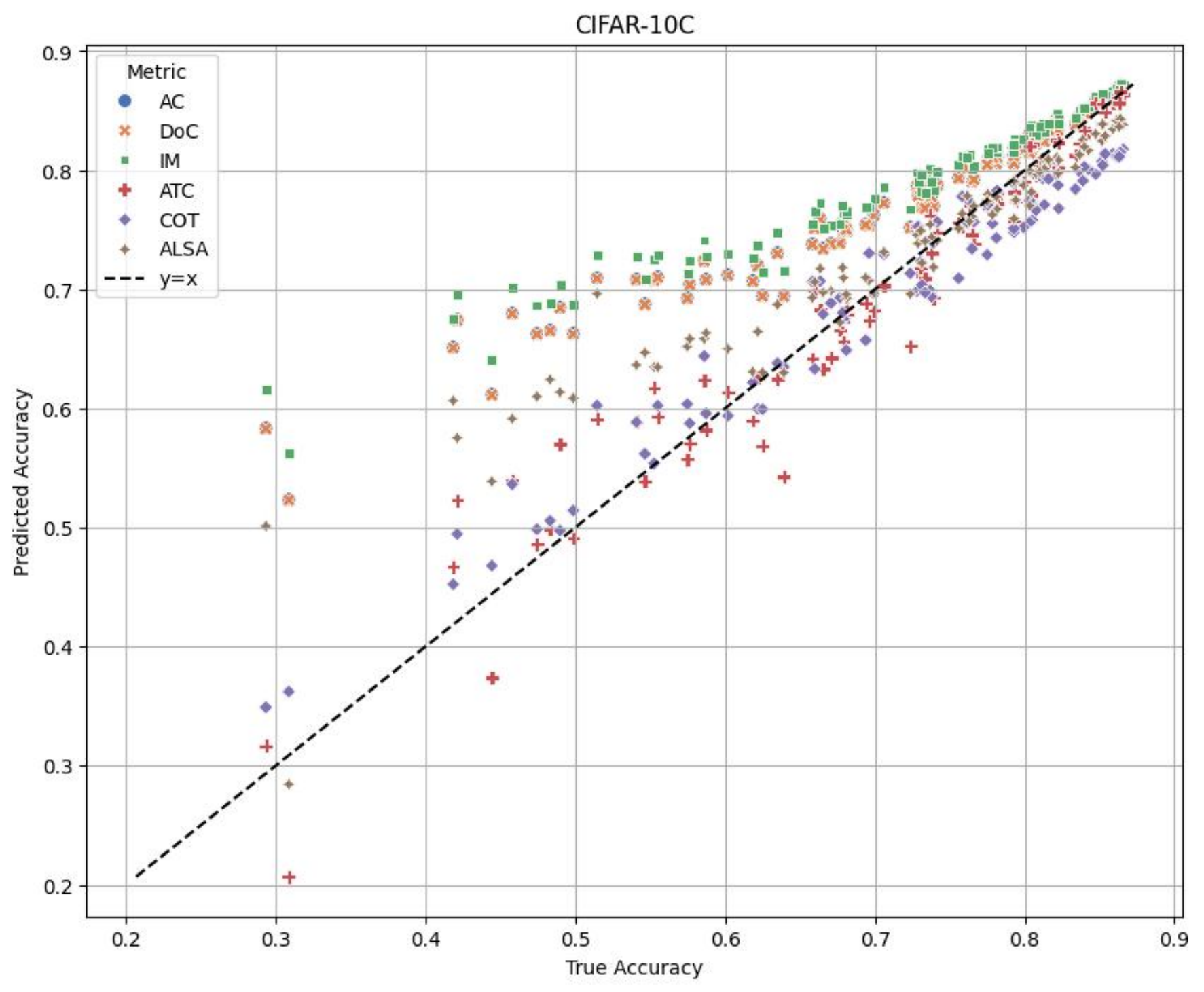}
} &
\bmvaHangBox{
    \includegraphics[width=0.45\textwidth]{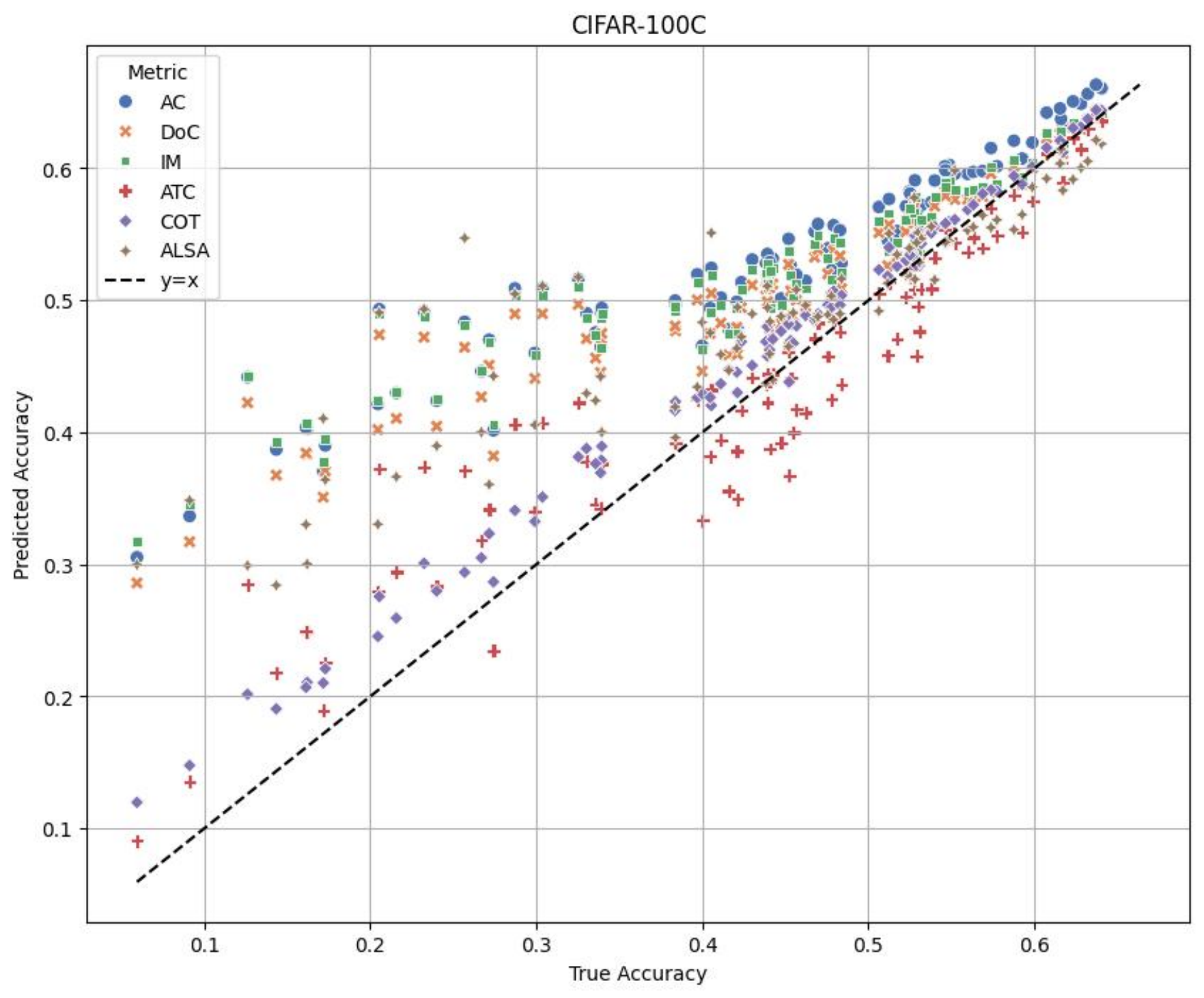}
} \\
(a) Predicted vs True on CIFAR-10C &
(b) Predicted vs True on CIFAR-100C
\end{tabular}
\caption{Visualization of evaluation results on CIFAR-10C and CIFAR-100C with ResNet18 models. The ResNet18 model trained on CIFAR-10 is evaluated on CIFAR-10C, while a separate model trained on CIFAR-100 is evaluated on CIFAR-100C.}
\label{fig:cifar_vis}
\end{figure*}

Figure~\ref{fig:cifar_vis} presents the evaluation results on CIFAR-10C (Figure~\ref{fig:cifar_vis}(a)) and CIFAR-100C (Figure~\ref{fig:cifar_vis}(b)), which contain controlled image corruptions that progressively degrade model performance. The proposed method assumes approximate alignment between the source and target conditional distributions $p(y|\mathbf{z})$, where $\mathbf{z} = f(\mathbf{x})$. This assumption generally holds under covariate shifts, but may become less valid when corruption levels are extremely severe.

As shown in the figure, ALSA delivers accurate estimates across a wide range of corruption intensities, maintaining high alignment with the ideal $y=x$ line even when the true accuracy drops by more than 30\%. Only under extreme corruption cases—where the predictive behavior of the model is heavily distorted—does the method exhibit a tendency to overestimate. Such behavior is expected, as accuracy estimation inherently relies on structural assumptions; in fact, it has been shown that no method can perform reliably under arbitrary distribution shifts~\cite{garg2022ATC}.

Despite these edge cases, ALSA demonstrates strong robustness in practice, outperforming baselines and providing reliable estimates across most realistic scenarios.

\subsection{ogbn-arvix Results Visualization}

\begin{figure*}[ht]
\centering
\begin{tabular}{cc}
\bmvaHangBox{
    \includegraphics[width=0.42\textwidth]{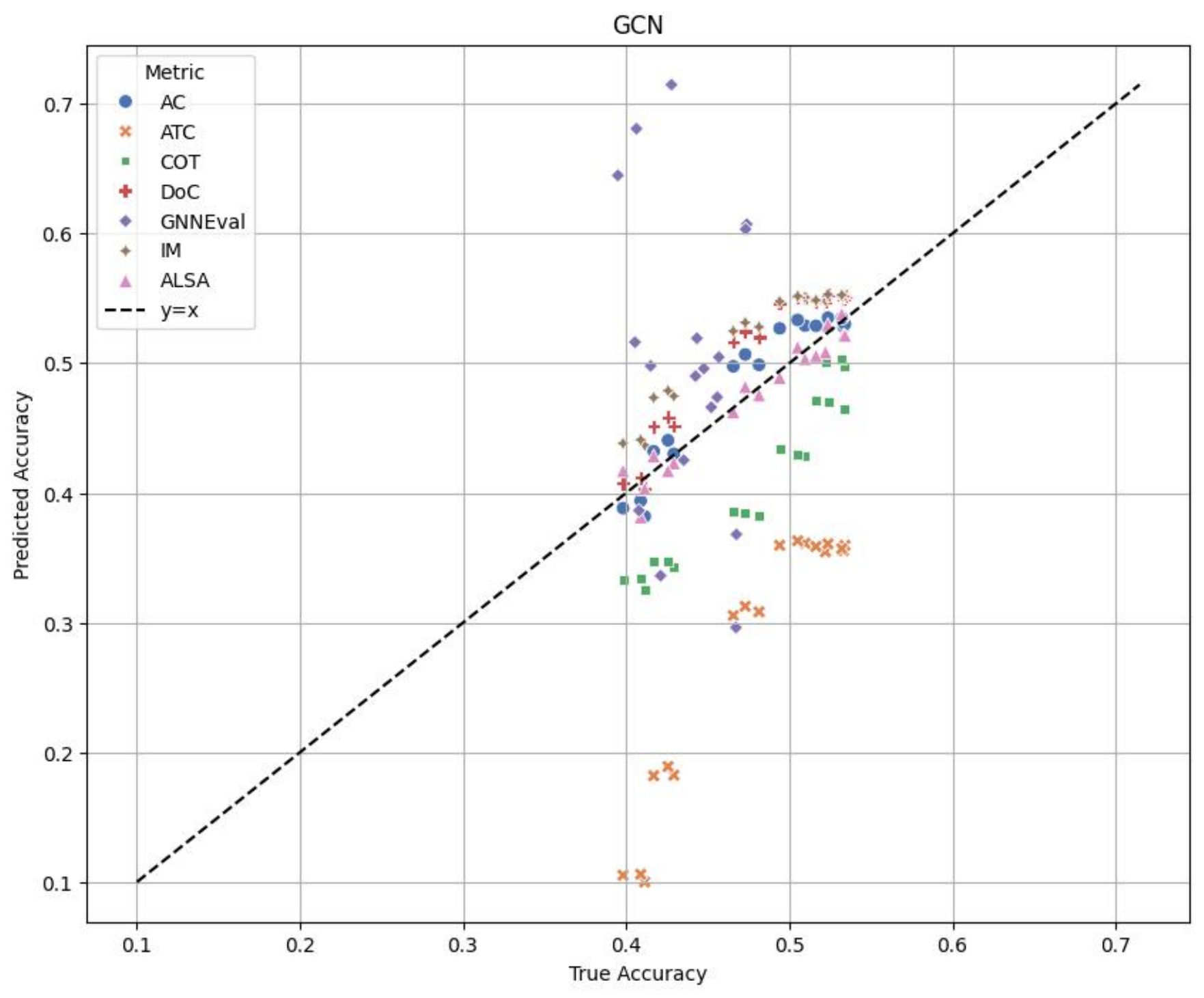}
} &
\bmvaHangBox{
    \includegraphics[width=0.42\textwidth]{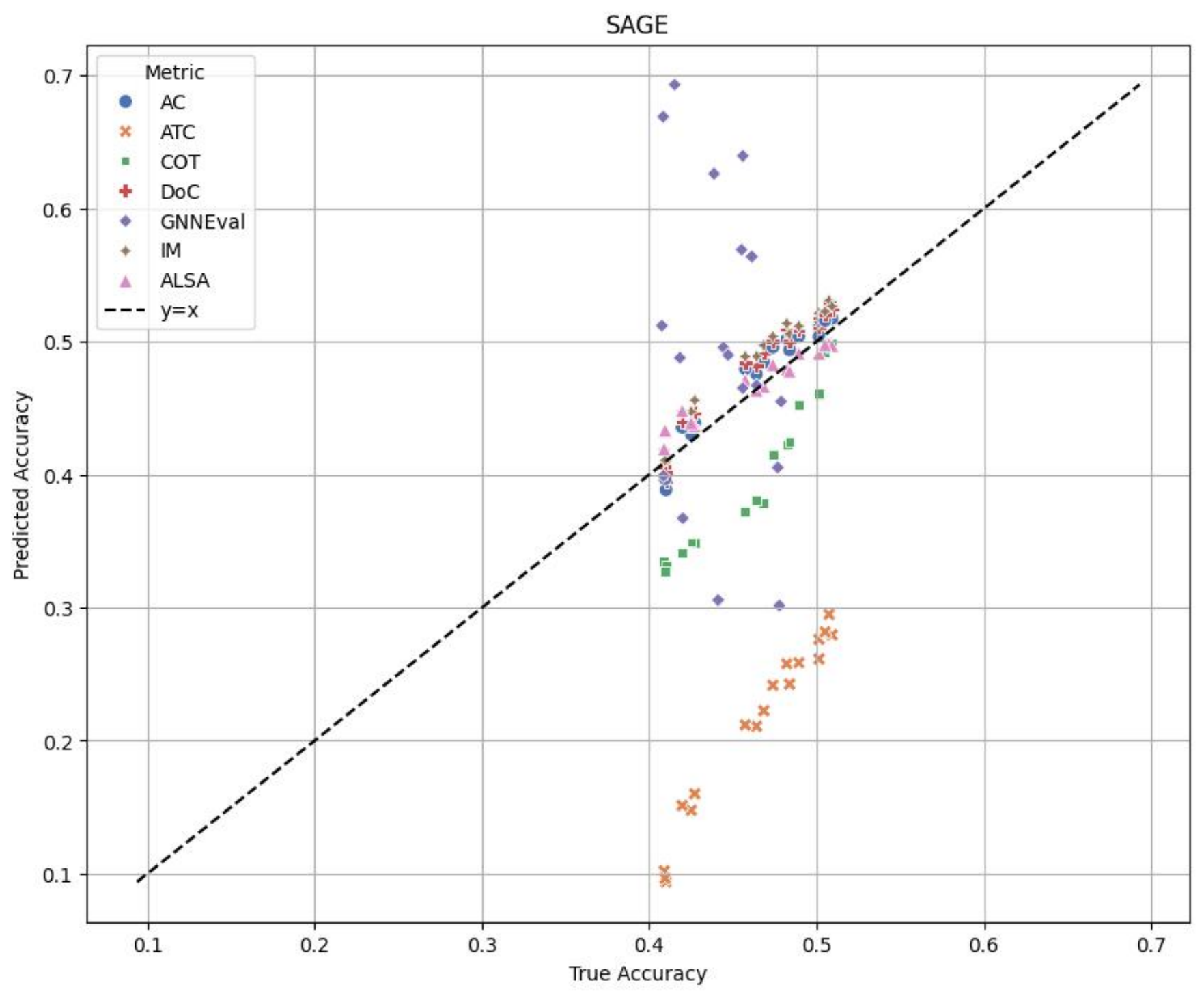}
} \\
(a) GCN Predicted vs True & (b) SAGE Predicted vs True \\
\\
\bmvaHangBox{
    \includegraphics[width=0.42\textwidth]{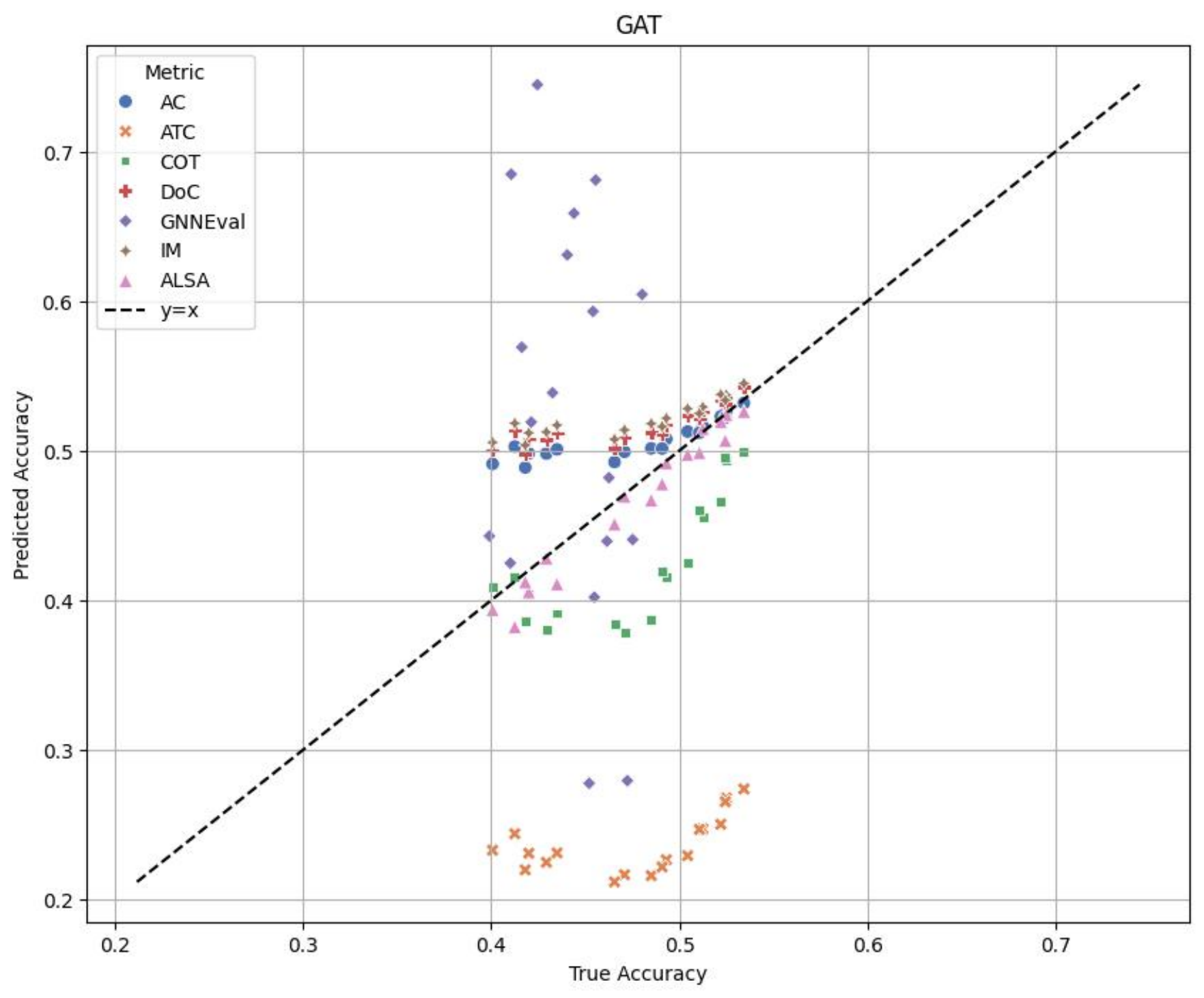}
} &
\bmvaHangBox{
    \includegraphics[width=0.42\textwidth]{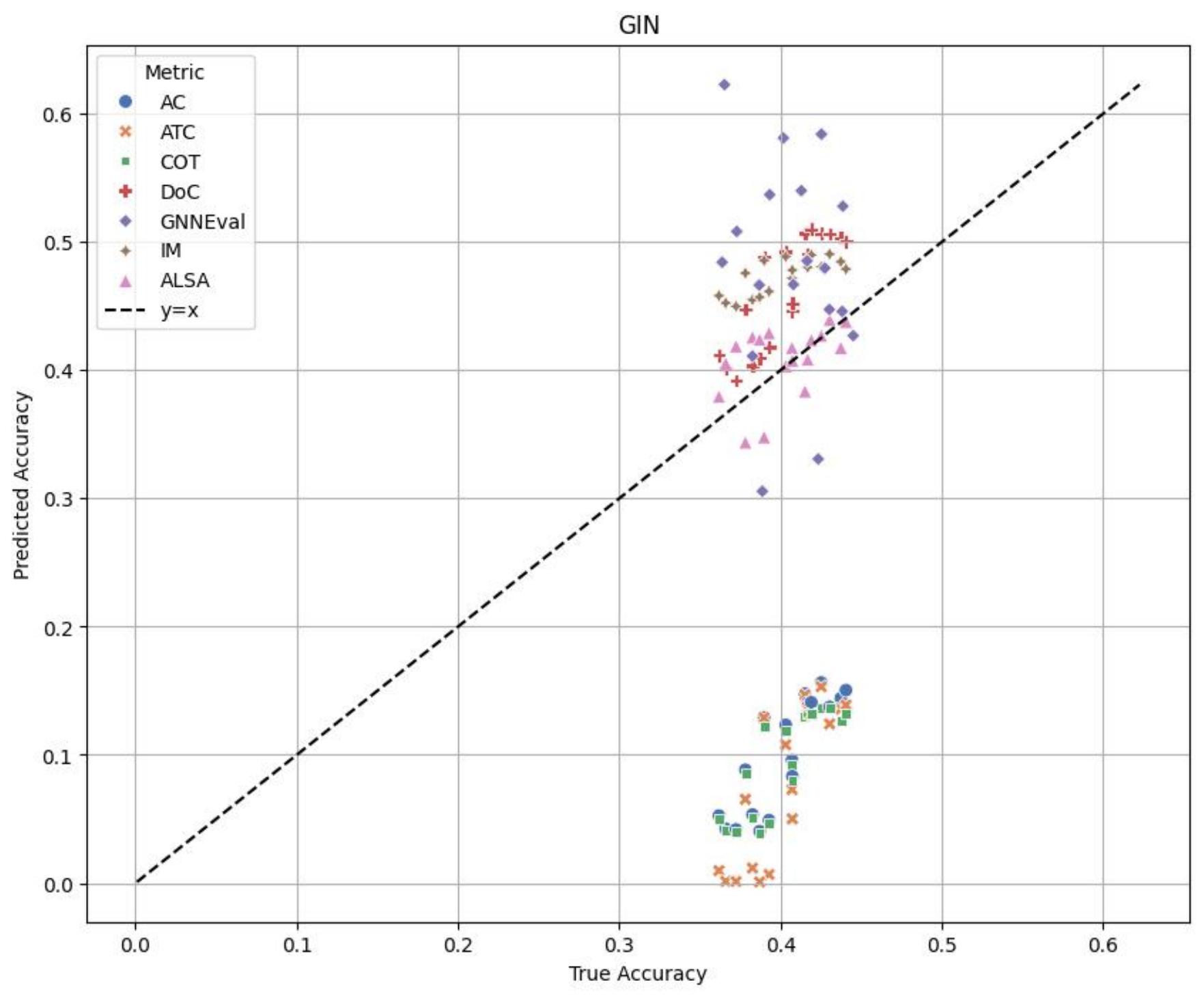}
} \\
(c) GAT Predicted vs True & (d) GIN Predicted vs True
\end{tabular}
\caption{Visualization of the evaluation results for four GNN models: GCN, GIN, GAT, and SAGE. The method is compared with all baseline methods.}
\label{fig:ogbn_arxiv}
\end{figure*}

In Figure~\ref{fig:ogbn_arxiv}, the evaluation results on four GNN models—GCN, GAT, GIN, and SAGE—are visualized. Temporal shifts are more likely to satisfy the covariate shift assumption, which validates the underlying assumption of the proposed method, even with significant changes in the marginal label distribution $p(y)$. As a result, it is observed that the method performs well across all models and outperforms all baselines. This is because the method can perceive differences in prediction performance across various classes.

\section{Hyperparameters Analysis}
\label{appendix:hyperparameters_analysis}

\subsection{Confidence Interval}
\begin{figure}[!hb]
\centering
\begin{tabular}{cc}
\bmvaHangBox{
    \includegraphics[width=0.45\columnwidth]{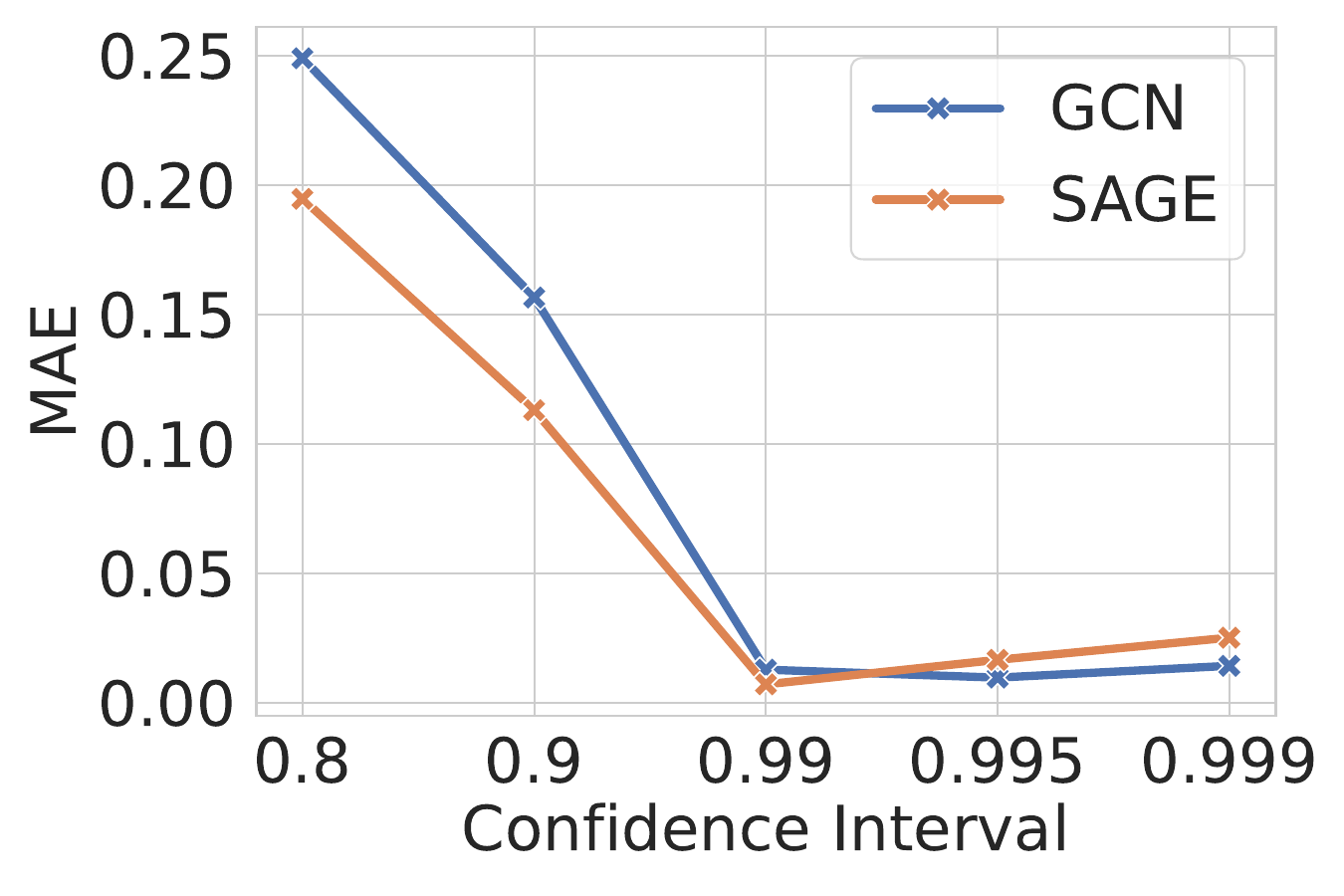}
} &
\bmvaHangBox{
    \includegraphics[width=0.45\columnwidth]{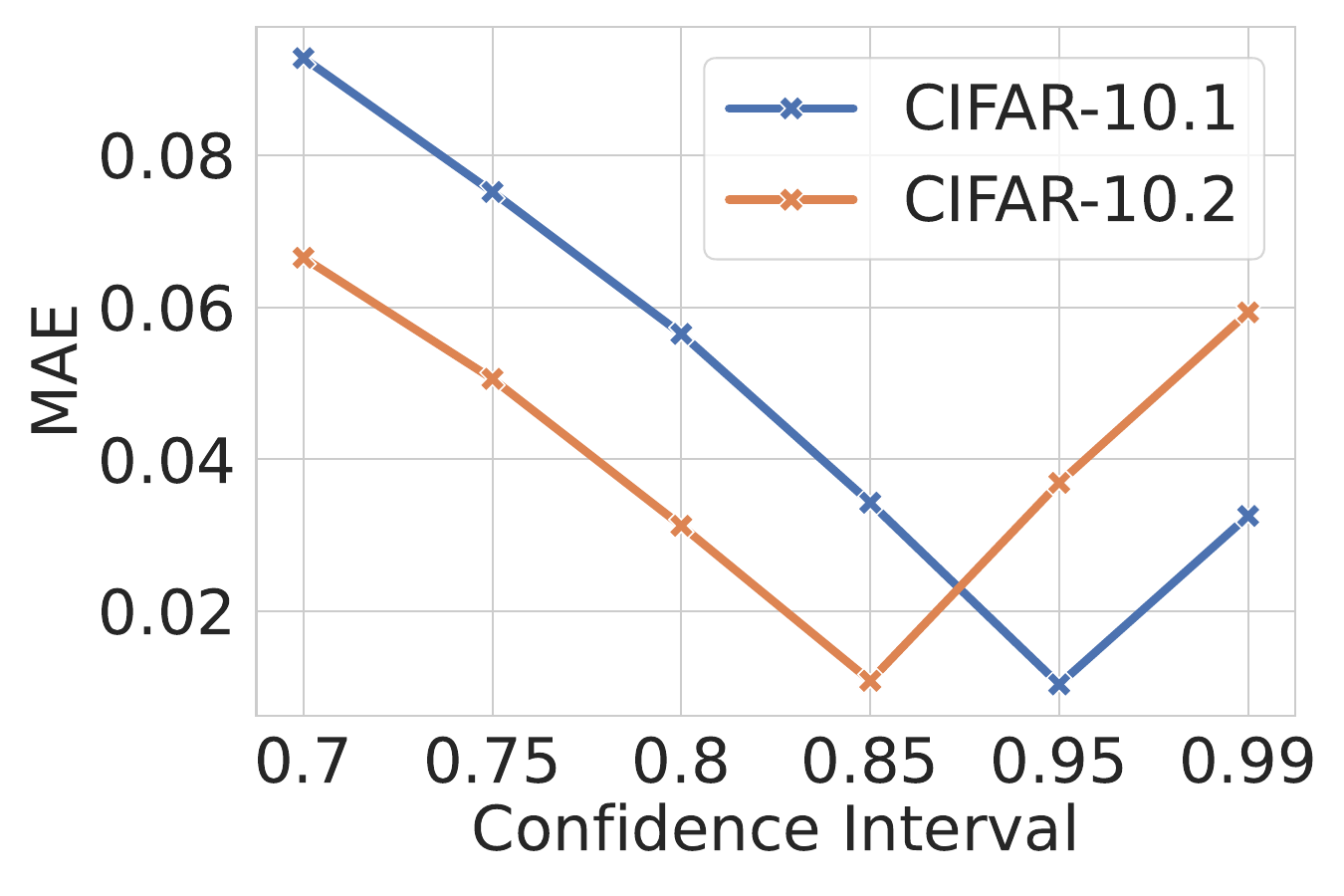}
} \\
(a) GCN and SAGE for ogbn-arxiv & (b) CIFAR-10.1 and CIFAR-10.2
\end{tabular}
\caption{MAE with increasing confidence interval $\alpha$.}
\label{fig:confidence_params_analysis}
\end{figure}

The confidence interval $\alpha$ is used to determine when to adjust the predicted probability of logit prediction correctness, particularly in case where no single anchor provides sufficient influence on a logit at a specific position in the logit space. Our method generalizes well for specific positions when $\alpha$ is small, however, the adjusted prediction will lead to an underestimation of accuracy and resulting in a high MAE. On the contrary, when $\alpha$ is too large, the method tends to avoid adjustments for positions where it is likely to fail in generalizing to a reliable prediction. In such cases, the negligible cumulative influence tends toward 0, producing an estimated accuracy of 0.5, which overestimates the actual accuracy and again increases the MAE. This trend is illustrated in Figure~\ref{fig:confidence_params_analysis}.

\subsection{Number of Anchors}

\begin{figure}[h]
\centering
\begin{tabular}{cc}
\bmvaHangBox{
    \includegraphics[width=0.45\columnwidth]{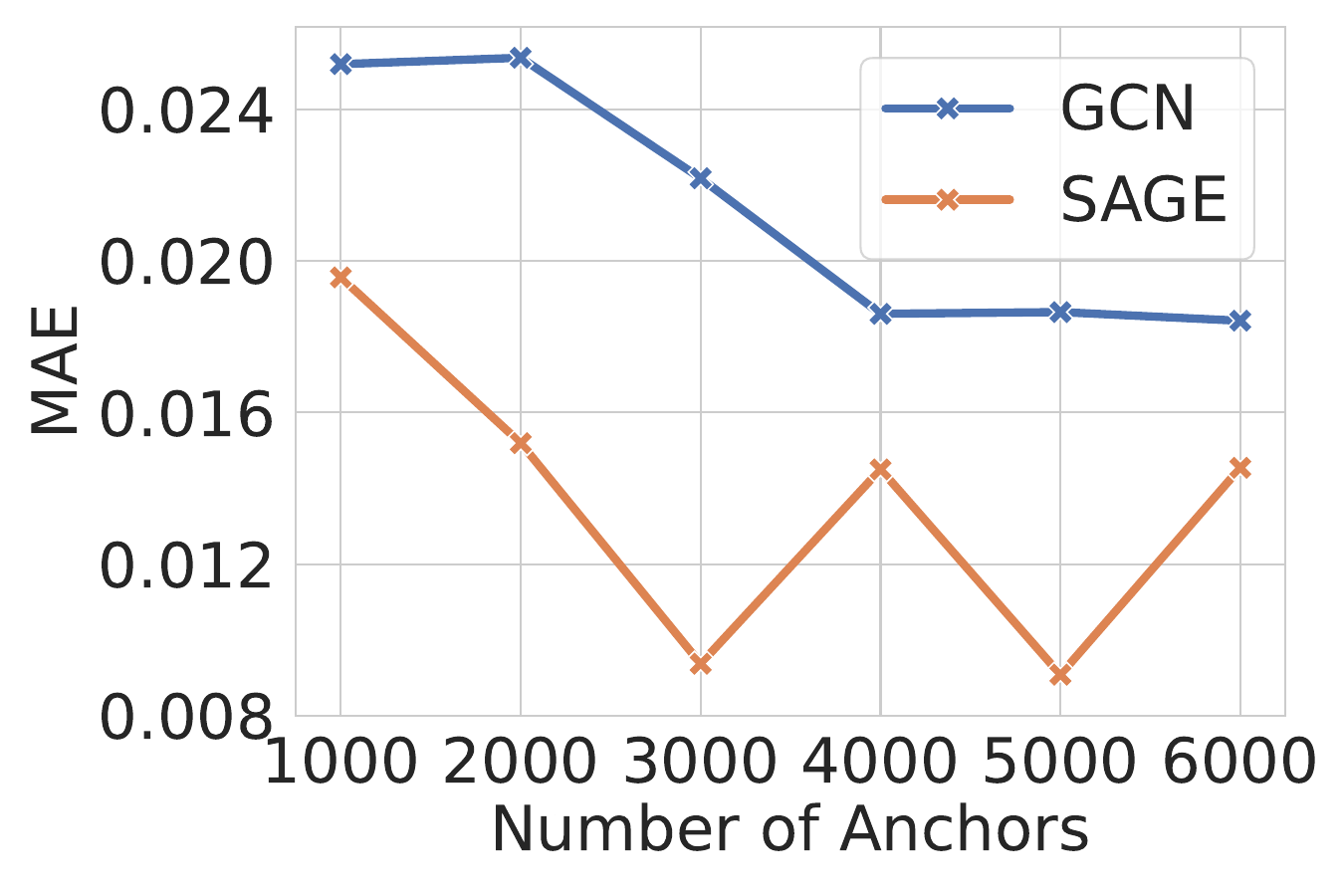}
} &
\bmvaHangBox{
    \includegraphics[width=0.45\columnwidth]{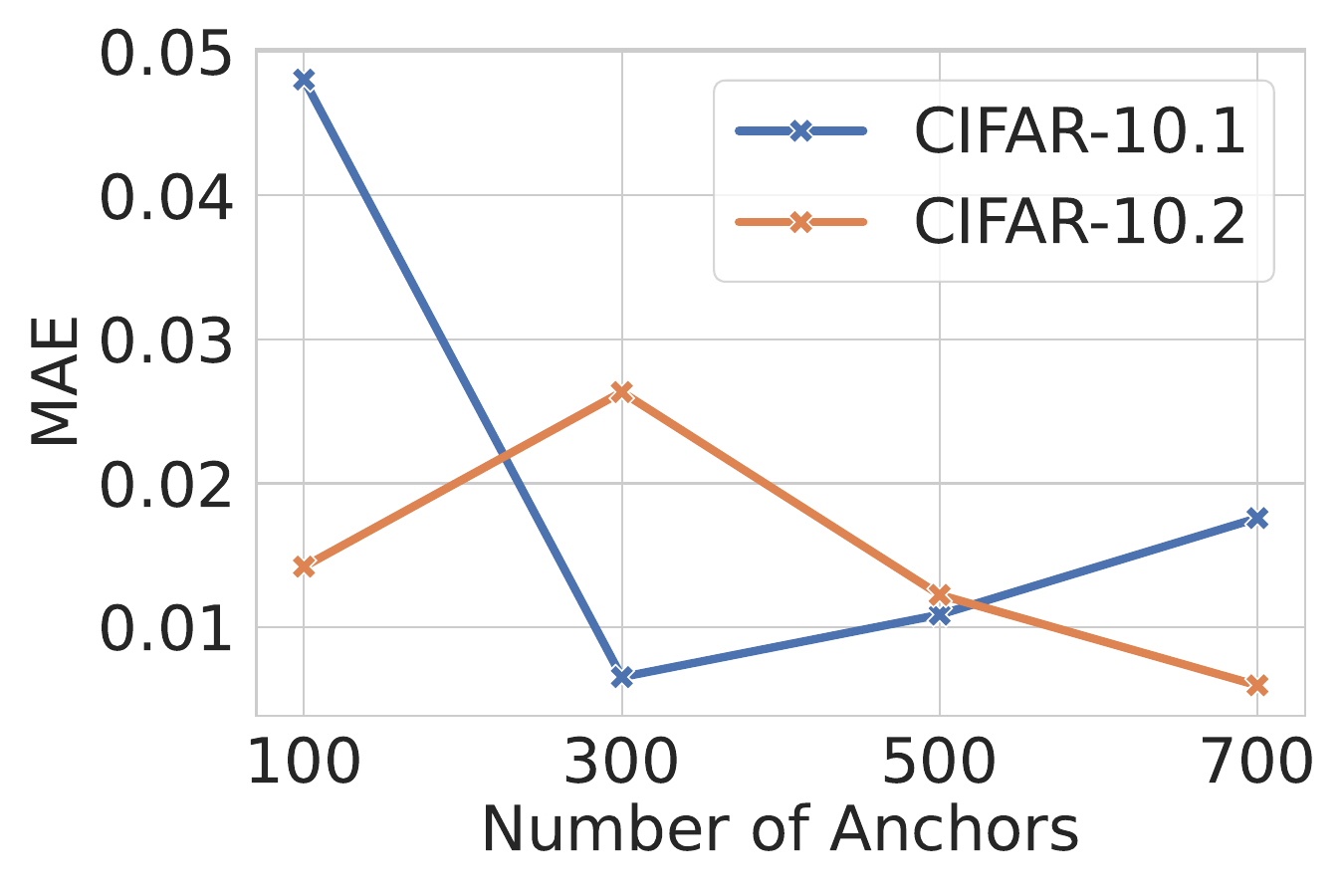}
} \\
(a) GCN and SAGE for ogbn-arxiv & (b) CIFAR-10.1 and CIFAR-10.2
\end{tabular}
\caption{MAE with increasing the number of anchors.}
\label{fig:anchor_size_params_analysis}
\end{figure}

When the number of anchors is too small, the model may struggle to provide sufficiently generalized accuracy, as the limited number of anchors makes it difficult to capture the probability distribution demonstrated by the model on the validation set. As the number of anchors increases, the MAE decreases rapidly, but once the number of anchors reaches a certain threshold, further increases yield no significant improvement. Such trend can be observed in Figure~\ref{fig:anchor_size_params_analysis}.

\section{Complexity Analysis}
\label{appendix:complexity_analysis}

\begin{wraptable}[11]{r}{0.5\textwidth}
\vspace{-8pt}
\centering
\renewcommand{\arraystretch}{1.1}
\setlength{\tabcolsep}{5pt}
\footnotesize
\begin{tabular}{lcc}
\toprule
\textbf{Dataset} & \textbf{Anchors} & \textbf{Training Time} \\
\midrule
CIFAR10       & 3000  & $\sim$10 sec \\
CIFAR100      & 3000  & $\sim$20 sec \\
ImageNet      & 20000 & $\sim$30 min \\
Amazon-WILD   & 5000  & $\sim$4 min  \\
Waterbirds    & 500   & $\sim$20 sec \\
MNIST         & 3000  & $\sim$30 sec \\
Office-31     & 300   & $\sim$5 sec  \\
\bottomrule
\end{tabular}
\vspace{2pt}
\caption{Number of anchors and training time for different datasets.}
\label{tab:anchor_training_time}
\vspace{-10pt}
\end{wraptable}

\begin{figure}[h]
\centering
\begin{tabular}{cc}
\bmvaHangBox{
    \includegraphics[width=0.45\columnwidth]{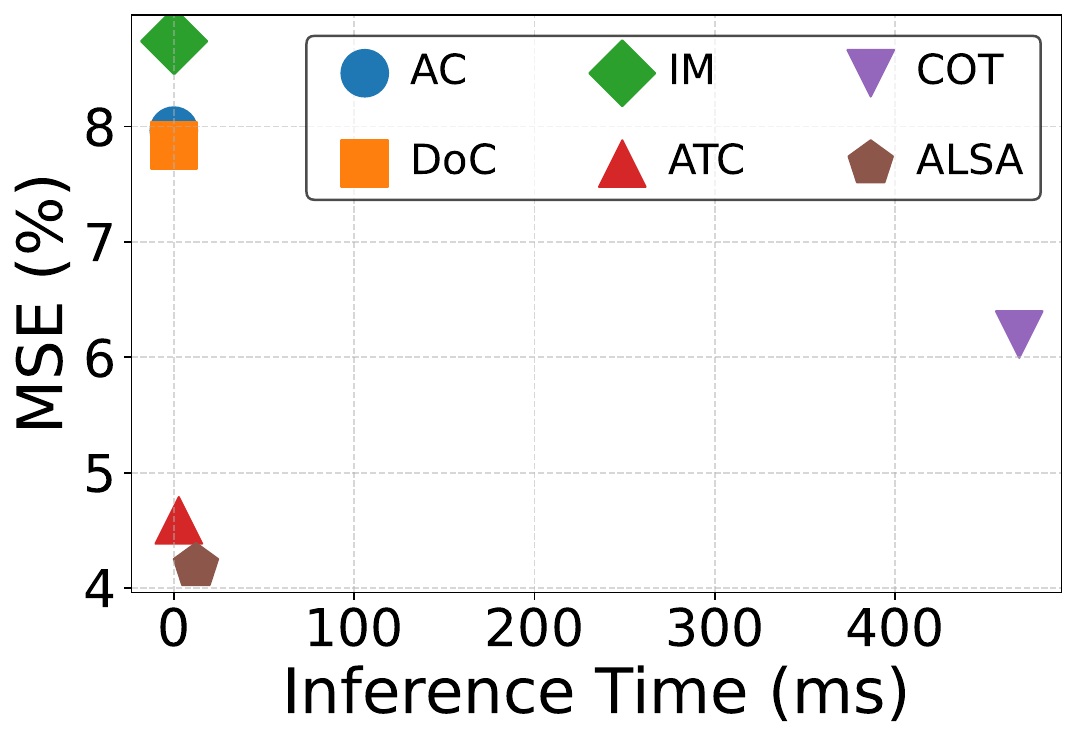}
} &
\bmvaHangBox{
    \includegraphics[width=0.45\columnwidth]{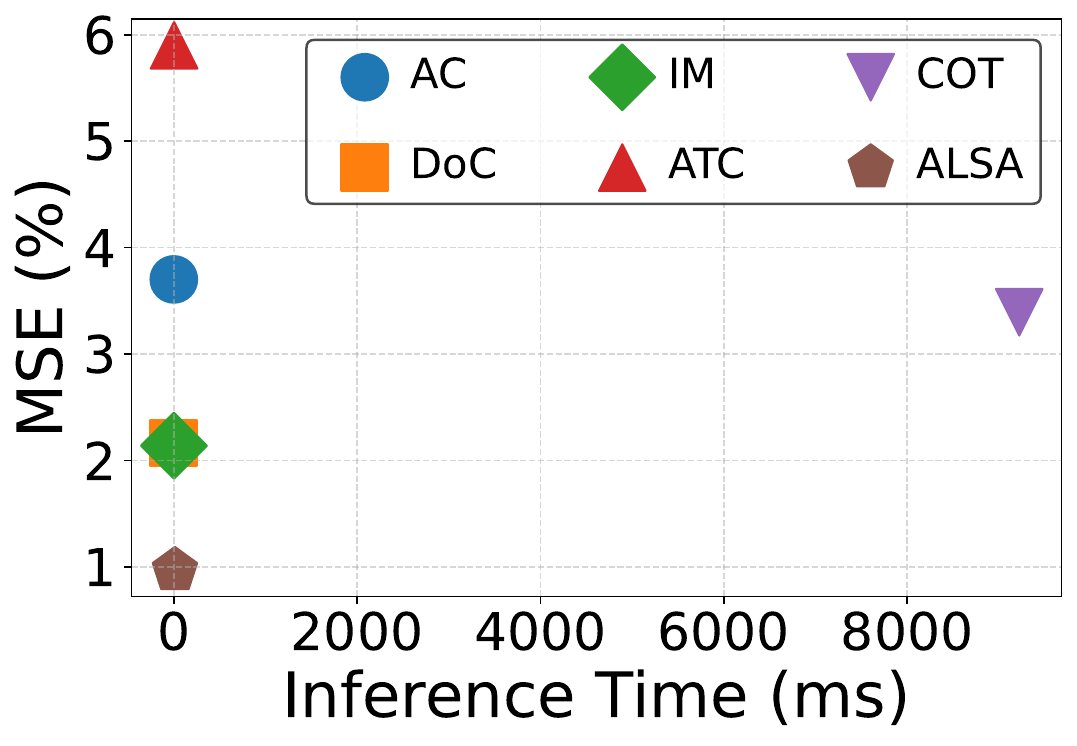}
} \\
(a) Inference on CIFAR10.2 & (b) Inference on Amazon-WILDS
\end{tabular}
\caption{Inference time and MAE across different methods.}
\label{fig:complexity_analysis}
\end{figure}

The model needs to compute the influence of each anchor and aggregate these influences to obtain the predicted probability, resulting in a complexity of $O(mn)$ in both the training and inference phases, where $n$ is the number of training or testing samples and $m$ is the number of anchors. Once $m$ is determined and considered as a constant, \textbf{the complexity in the inference phase is linear}, enabling scalability to large datasets while maintaining the benefits of fast inference. The comparison of inference time on CIFAR-10.2 and Amazon-WILDS is reported in Figure~\ref{fig:complexity_analysis}(a) and Figure~\ref{fig:complexity_analysis}(b), respectively. ALSA achieves a well-balanced time complexity and estimation accuracy. The training time on each dataset is also reported in Table~\ref{tab:anchor_training_time}.

\end{document}